\documentclass[journal]{IEEEtran}
\pdfoutput=1
\ifCLASSINFOpdf
\else
\fi

\usepackage{amsmath}
\usepackage{algpseudocode}

\usepackage{hyperref}
\usepackage{graphicx}
\usepackage{booktabs}
\newcommand{\ra}[1]{\renewcommand{\arraystretch}{#1}}
\usepackage{adjustbox}
\usepackage{subfig}

\usepackage[ruled,vlined]{algorithm2e}

\SetKwInput{kwTerms}{Terms}

\SetKwInput{kwTrain}{Train the Encoder / Decoder}

\SetKwInput{kwGen}{Generate Samples}
\SetKwInput{kwForm}{Formulas}

\usepackage{tikz}

\usetikzlibrary{arrows.meta,
                chains,
                positioning,
                shapes.geometric
                }

\usepackage{lipsum,multicol}

\begin{document}
%
\title{DeepSMOTE: Fusing Deep Learning and SMOTE for Imbalanced Data}

\author{Damien~Dablain,
        Bartosz~Krawczyk$^{\dagger}$,~\IEEEmembership{Member,~IEEE,}
        and~Nitesh~V.~Chawla,~\IEEEmembership{Senior Member,~IEEE,}
\thanks{$^{\dagger}$ corresponding author}
\thanks{D. Dablain and N.V. Chawla are with the Department of Computer Science and Engineering, and the Interdisciplinary Center for Network Science and Applications (iCeNSA), the University of Notre Dame, Notre Dame, IN 46556 e-mail: \{ddablain,nchawla\}@nd.edu}
\thanks{B. Krawczyk is with Department of Computer Science, Virginia Commonwealth University, Richmond, VA 23284 e-mail: bkrawczyk@vcu.edu}
}


\maketitle

\begin{abstract}
Despite over two decades of progress, imbalanced data is still considered a significant challenge for contemporary machine learning models. Modern advances in deep learning have magnified the importance of the imbalanced data problem. The two main approaches to address this issue are based on loss function modifications and instance resampling. Instance sampling is typically based on Generative Adversarial Networks (GANs), which may suffer from mode collapse. Therefore, there is a need for an oversampling method that is specifically tailored to deep learning models, can work on raw images while preserving their properties, and is capable of generating high quality, artificial images that can enhance minority classes and balance the training set. We propose DeepSMOTE - a novel oversampling algorithm for deep learning models. It is simple, yet effective in its design. It consists of three major components: (i) an encoder/decoder framework; (ii) SMOTE-based oversampling; and (iii) a dedicated loss function that is enhanced with a penalty term.  An important advantage of DeepSMOTE over GAN-based oversampling is that DeepSMOTE does not require a discriminator, and it generates high-quality artificial images that are both information-rich and suitable for visual inspection. DeepSMOTE code is publicly available at: \href{https://github.com/dd1github/DeepSMOTE}{https://github.com/dd1github/DeepSMOTE}.
\end{abstract}


\IEEEpeerreviewmaketitle

\section{Introduction}
\label{sec:int}

\IEEEPARstart{L}{earning} from imbalanced data is among the crucial problems faced by the machine learning community \cite{krawczyk2016learning}. Imbalanced class distributions affect the training process of classifiers, leading to unfavourable bias towards the majority class(es). This may result in high error, or even complete omission, of the minority class(es). Such a situation cannot be accepted in most real-world applications (e.g., medicine or intrusion detection) and thus algorithms for countering the class imbalance problem have been a focus of intense research for over two decades \cite{Fernandez:2018}. Contemporary applications have extended our view of the problem of imbalanced data, confirming that disproportionate classes are not the sole source of learning problems. A skewed class imbalance ratio is often accompanied by additional factors, such as difficult and borderline instances, small disjuncts, small sample size, or the drifting nature of streaming data \cite{Fernandez:2018}. These continuously emerging challenges keep the field expanding, calling for novel and effective solutions that can analyze, understand, and tackle these data-level difficulties. Deep learning is currently considered as the most promising branch of machine learning, capable of achieving outstanding cognitive and recognition potentials. However, despite its powerful capabilities, deep architectures are still very vulnerable to imbalanced data distributions \cite{Bao:2020,Bugnon:2020} and are affected by novel challenges such as complex data representations \cite{Jing:2021}, the relationship between imbalanced data and extracted embeddings \cite{Wang:2018}, and learning from an extremely large number of classes \cite{Wu:2020}. 

\noindent \textbf{Research goal.} We propose a novel oversampling method for imbalanced data that is specifically tailored to deep learning models and that leverages the advantages of SMOTE \cite{chawla2002smote}, while embedding it in a deep architecture capable of efficient operation on complex data representations, such as images. 

\smallskip
\noindent \textbf{Motivation.} Although the imbalanced data problem strongly affects both deep learning models \cite{Huang:2020} and their shallow counterparts, there has been limited research on how to counter this challenge in the deep learning realm. In the past, the two main directions that have been pursued to overcome this challenge have been loss function modifications and resampling approaches. The deep learning resampling solutions are either pixel-based or use GANs for artificial instance generation. Both of these approaches suffer from strong limitations. Pixel-based solutions often cannot capture complex data properties of images and are not capable of generating meaningful artificial images. GAN-based solutions require significant amounts of data, are difficult to tune, and may suffer from mode collapse \cite{miyato2018spectral, salimans2016improved,
 gulrajani2017improved, arjovsky2017wasserstein}. Therefore, there is a need for a novel oversampling method that is specifically tailored to the nature of deep learning models, can work on raw images while preserving their properties, and is capable of generating artificial images that are of both of high visual quality and enrich the discriminative capabilities of deep models. 

\smallskip
\noindent \textbf{Summary.} We propose DeepSMOTE - a novel oversampling algorithm for deep learning models based on the highly popular SMOTE method. Our method bridges the advantages of metric-based resampling approaches that use data characteristics to leverage their performance, with a deep architecture capable of working with complex and high-dimensional data. DeepSMOTE consists of three major components: (i) an encoder/decoder framework; (ii) SMOTE-based oversampling; and (iii) a dedicated loss function enhanced with a penalty term. This approach allows us to embed effective SMOTE-based artificial instance generation within a deep encoder / decoder model for a streamlined and end-to-end process, including low dimensional embeddings, artificial image generation, and multi-class classification.

\smallskip
\noindent \textbf{Main contributions.}  In order for an oversampling method to be successfully applied to deep learning models, we believe that it should meet three essential criteria: (1) it should operate in an end-to-end manner, (2) it should learn a representation of the raw data and embed the data into a lower dimensional \emph{feature space}, and (3) it should readily generate output (e.g., images) that can be visually inspected.  In this paper, we propose DeepSMOTE, which meets these three criteria, and also offer the following scientific contributions to the field of deep learning under class imbalance:

\begin{itemize}
    \item \textbf{Deep oversampling architecture.} We introduce DeepSMOTE - a self-contained deep architecture for oversampling and artificial instance generation that allows efficient handling of complex-imbalanced and high-dimensional data, such as images.
    
    \item \textbf{Simple and effective solution to class imbalance.} Our framework is simple, yet effective in its design. It consists of only three major components responsible for low dimensional representations of raw data, resampling, and classification. 
    
    \item \textbf{No need for a discriminator during training.} An important advantage of DeepSMOTE over GAN-based oversampling lies in the fact that DeepSMOTE does not require a discriminator during the artificial instance generation process. We propose a penalty function that ensures efficient usage of training data to prime our generator. 
    
    \item \textbf{High quality image generation.} DeepSMOTE generates high-quality artificial images that are both suitable for visual inspection (they are of identical quality as their real counterparts), and information-rich, which allows for efficient balancing of classes and alleviates the effects of imbalanced distributions. 
    
    \item \textbf{Extensive experimental study.} We propose a carefully designed and thorough experimental study that compares DeepSMOTE with state-of-the-art oversampling and GAN-based methods. Using five popular image benchmarks and three dedicated skew-insensitive metrics over two different testing protocols, we empirically prove the merits of DeepSMOTE over the reference algorithms. Furthermore, we show that DeepSMOTE displays an excellent robustness to increasing imbalance ratios, being able to efficiently handle even extremely skewed problems.

\end{itemize}

\smallskip
\noindent \textbf{Paper outline.} In this paper, we first provide an overview of the imbalanced data problem and the traditional approaches that have been employed to overcome this issue.  Next, we discuss how deep learning methods have been  used to generate data and augment imbalanced datasets.  We then introduce our approach to imbalanced learning, which combines deep learning with SMOTE.  Finally, we discuss our extensive experimentation, which validates the benefits of DeepSMOTE.

\section{Learning from imbalanced data}
\label{sec:imb}

The first works on imbalanced data came from binary classification problems. Here, the presence of majority and minority classes is assumed, with a specific imbalance ratio. Such skewed class distributions pose a challenge for machine learning models, as standard classifiers are driven by a 0-1 loss function that assumes a uniform penalty over both classes. Therefore, any learning procedure driven by such a function will lead to a bias towards the majority class. At the same time, the minority class is usually more important and thus cannot be poorly recognized. Therefore, methods dedicated to overcoming the imbalance problem aim at either alleviating the class skew or alternating the learning procedure. The three main approaches are:

\noindent \textbf{Data-level approaches.} This solution should be viewed as a preprocessing phase that is classifier-independent. Here, we focus on balancing the dataset before applying any classifier training. This is usually achieved in one of three ways: (i) reducing the size of the majority class (undersampling); (ii) increasing the size of minority class (oversampling); or (iii) a combination of the two previous solutions (hybrid approach). Both under- and oversampling can be performed in a random manner, which has low complexity, but leads to potentially unstable behavior (e.g., removing important instances or enhancing noisy ones). Therefore, guided solutions have been proposed that try to smartly choose instances for preprocessing. While not many solutions have been proposed for guided undersampling \cite{Koziarski:2020,Lin:2017,Vuttipittayamongkol:2020}, oversampling has gained much more attention due to the success of SMOTE \cite{chawla2002smote}, which led to the introduction of a plethora of variants \cite{Douzas:2019,He:2008,Liang:2020}. However, recent works show that SMOTE-based methods cannot properly deal with multi-modal data and cases with high intra-class overlap or noise. Therefore, completely new approaches that do not rely on $k$-nearest neighbors have been successfully developed \cite{Koziarski:2019,Koziarski:2017}.
 
\noindent \textbf{Algorithm-level approaches.} Contrary to the previously discussed approaches, algorithm-level solutions work directly within the training procedure of the considered classifier. Therefore, they lack the flexibility offered by data-level approaches, but compensate with a more direct and powerful way of reducing the bias of the learning algorithm. They also require an in-depth understanding of how a given training procedure is conducted and what specific part of it may lead to bias towards the majority class. The most commonly addressed issues with the algorithmic approach are developing novel skew-insensitive split criteria for decision trees \cite{Boonchua:2017,Cieslak:2012,Li:2018}, using instance weighting for Support Vector Machines \cite{Datta:2019,Fan:2017,Qi:2019}, or modifying the way different layers are trained in deep learning \cite{Dong:2019,Liu:2018icdm,Wang:2019}. Furthermore, cost-sensitive solutions \cite{Cao:2018,Khan:2018,Zhang:2019} and one-class classification \cite{Devi:2019,Krawczyk:2014,Perez-Sanchez:2015} can also be considered as a form of algorithm-level approaches. 
 
\noindent \textbf{Ensemble approaches.} The third way of managing imbalanced data is to use ensemble learning \cite{Wozniak:2014}. Here, one either combines a popular ensemble architecture (usually based on Bagging or Boosting) with one of the two previously discussed approaches, or develops a completely new ensemble architecture that is skew-insensitive on its own \cite{Diez-Pastor:2015kbs}. One of the most successful families of methods is the combination of Bagging with undersampling \cite{Blaszczynski:2015,Hido:2009,Roshan:2020}, Boosting with any resampling technique \cite{Datta:2020,Krawczyk:2016asc,Zhang:2018b}, or cost-sensitive learning with multiple classifiers \cite{Krawczyk:2014asc,Tao:2019,Zhou:2016}. Data-level techniques can be used to manage the diversity of the ensemble \cite{Diez-Pastor:2015}, which is a crucial factor behind the predictive power of multiple classifier systems. Additionally, to manage the individual accuracy of classifiers and eliminate weaker learners, one may use dynamic classifier selection \cite{Roy:2018} and dynamic ensemble selection \cite{Zyblewski:2021}, which ensures that the final decision will be based only on the most competent classifiers from the pool \cite{Souza:2019}.

\section{Deep learning from imbalanced data}
\label{sec:dim}

Since the imbalanced data problem has been attracting increasing attention from the deep learning community, let us discuss three main trends in this area.

\smallskip
\noindent \textbf{Instance generation with deep neural networks.} Recent works that combine deep learning with shallow oversampoling methods do not give desirable results and traditional resampling approaches cannot efficiently augment the training set for deep models  \cite{Fernandez:2018,Bellinger:2020}. This leads to an interest in generative models and adapting them to work in a similar manner to oversampling techniques \cite{Fajardo:2021}. An encoder / decoder combination can efficiently introduce artificial instances into a given embedding space \cite{Bellinger:2018}. Generative Adversarial Networks (GAN)  \cite{goodfellow2014generative}, 
Variational Autoencoders (VAE) \cite{kingma2013auto}, and Wasserstein 
Autoencoders (WAE) \cite{tolstikhin2017wasserstein} have been successfully
 used within computer vision \cite{zhu2017unpaired, karras2020analyzing} and
 robotic control \cite{watter2015embed, bonatti2019learning} to learn the
 latent distribution of data. These techniques can also be extended to data generation for oversampling
 (e.g., medical imaging) \cite{yi2019generative}.

VAEs operate by maximizing a variational lower bound of the data log-likelihood
 \cite{hu2017unifying, doersch2016tutorial}.  The
 loss function in a VAE is typically implemented by combining a reconstruction
 loss with the Kullback-Leibler (KL) divergence.  The KL divergence can be 
interpreted as an implicit penalty on the reconstruction loss.  By penalizing
 the reconstruction loss, the model can learn to \emph{vary} its reconstruction
 of the data distribution and thus \emph{generate} output (e.g., images)
 based on a latent distribution of the input.  

WAEs also exhibit generative qualities.  Similar to VAEs, the loss function
 of a WAE is often implemented by combining a reconstruction loss with a
 penalty term.  In the case of a WAE, the penalty term is expressed as the
 output of a discriminator network.  

GANs have achieved impressive results in the computer vision arena
 \cite{wu2019logan, chen2016infogan}.  GANs formulate image generation as
 a min-max game between a generator and a discriminator network
 \cite{pfau2016connecting}.  Despite their impressive results, GANs
 require the use of two networks, are sometimes difficult to train and 
are subject to mode collapse (i.e., the repetitive generation of similar
 examples) \cite{miyato2018spectral, salimans2016improved,
 gulrajani2017improved, arjovsky2017wasserstein}.  

\noindent \textbf{Loss function adaptation.} One of the most popular approaches for making neural networks skew-insensitive is to modify their loss function. This approach successfully carried over to deep architectures and can be seen as an  algorithm-level modification. The idea behind modifying the loss function is based on the assumption that instances should not be treated uniformly during training and that errors on minority classes should be penalized more strongly, making it parallel to cost-sensitive learning \cite{Zhang:2019}. Mean False Error \cite{Wang:2016} and Focal Loss \cite{Lin:2017fl} are two of the most popular approaches based on this principle. The former simply balances the impact of instances from minority and majority classes, while the latter reduces the impact of easy instances on the loss function. More recently, multiple other loss functions were proposed, such as Log Bilinear Loss \cite{Resheff:2017}, Cross Entropy Loss \cite{Zhang:2018}, and Class-Balanced Loss \cite{Cui:2019}. 

\noindent \textbf{Long-tailed recognition.} This sub-field of deep learning evolved from problems where there is a high number of very rare classes that should nevertheless be properly recognized, despite their low sample size. Long-tailed recognition can be thus seen as an extreme case of the multi-class imbalanced problem, where we deal with a very high number of classes (hundreds) and an extremely high imbalance ratio. Due to very disproportionate class sizes, direct resampling is not advisable, as it will either significantly reduce the size of majority classes or require creation of too many artificial instances. Furthermore, classifiers need to handle the problem of small sample size, making learning from the tail classes very challenging. It is important to note that the majority of works in this domain assume that the test set is balanced. Very interesting solutions to this problem are based on adaptation of the loss function in deep neural networks, such as equalization loss \cite{Tan:2020}, hubless loss \cite{Abdelkarim:2020}, and range loss \cite{Zhang:2017}. Recent works suggest looking closer at class distributions and decomposing them into balanced sets -- an approach popular in traditional imbalanced classification. Zhou et al. \cite{Zhou:2020} proposed a cumulative learning scheme from global data properties down to class-based features. Sharma et al. \cite{Sharma:2020} suggests using a small ensemble of three classifiers, each focusing on majority, middle, or tail groups of classes. Meta-learning is also commonly used to improve the distribution estimation of tail classes \cite{Jamal:2020}.

\section{DeepSMOTE}
\label{sec:dsm}
\subsection{Motivation}

 We propose DeepSMOTE - a novel and breakthrough oversampling algorithm dedicated to enhancing deep learning models and countering the learning bias caused by imbalanced classes. As discussed above, oversampling is a proven technique for combating class imbalance; however, it has
 traditionally been used with classical machine learning models.  Several attempts
 have been made to extend oversampling methods, such as SMOTE, to deep learning models,
 although the results have been mixed \cite{ando2017deep, fernandez2018smote,
 johnson2019survey}.  Metric based oversampling methods, such as SMOTE, can also
 be computationally expensive because they require access to
 the full dataset during training and inference.  Accessing the full dataset,
 especially when dealing with image or speech data, can be challenging when using
 deep learning systems that also require large amounts of memory to store gradients.
 
 In order for an oversampling method to be successfully applied to deep
 learning models, we believe that it should meet three essential criteria:
 \begin{enumerate}
\item It should operate in an end-to-end manner by accepting raw input,
 such as images (i.e., similar to VAEs, WAEs and GANs). 
 \item It should learn a representation of the raw data and embed the data into a lower 
dimensional \emph{feature space}, which can be used for oversampling. 
\item It should readily generate output (e.g., images) that can be 
visually inspected, without extensive manipulation. 
\end{enumerate}

We show through our design steps and experimental evaluation that DeepSMOTE meets these criteria. In addition, it is capable of generating high-quality, sharp, and information-rich images without the need for a discriminator network. 

\subsection{Deep SMOTE Description}

DeepSMOTE consists of an encoder
 / decoder framework, a SMOTE-based oversampling method, and
 a loss function with a reconstruction loss and a penalty term.
Each of these features is discussed below, with Figure~1 depicting the flow of the DeepSMOTE approach, while the pseudo-code overview of DeepSMOTE is presented in Algorithm~1.

\begin{figure}[t!]
  \includegraphics[width=0.5\textwidth]{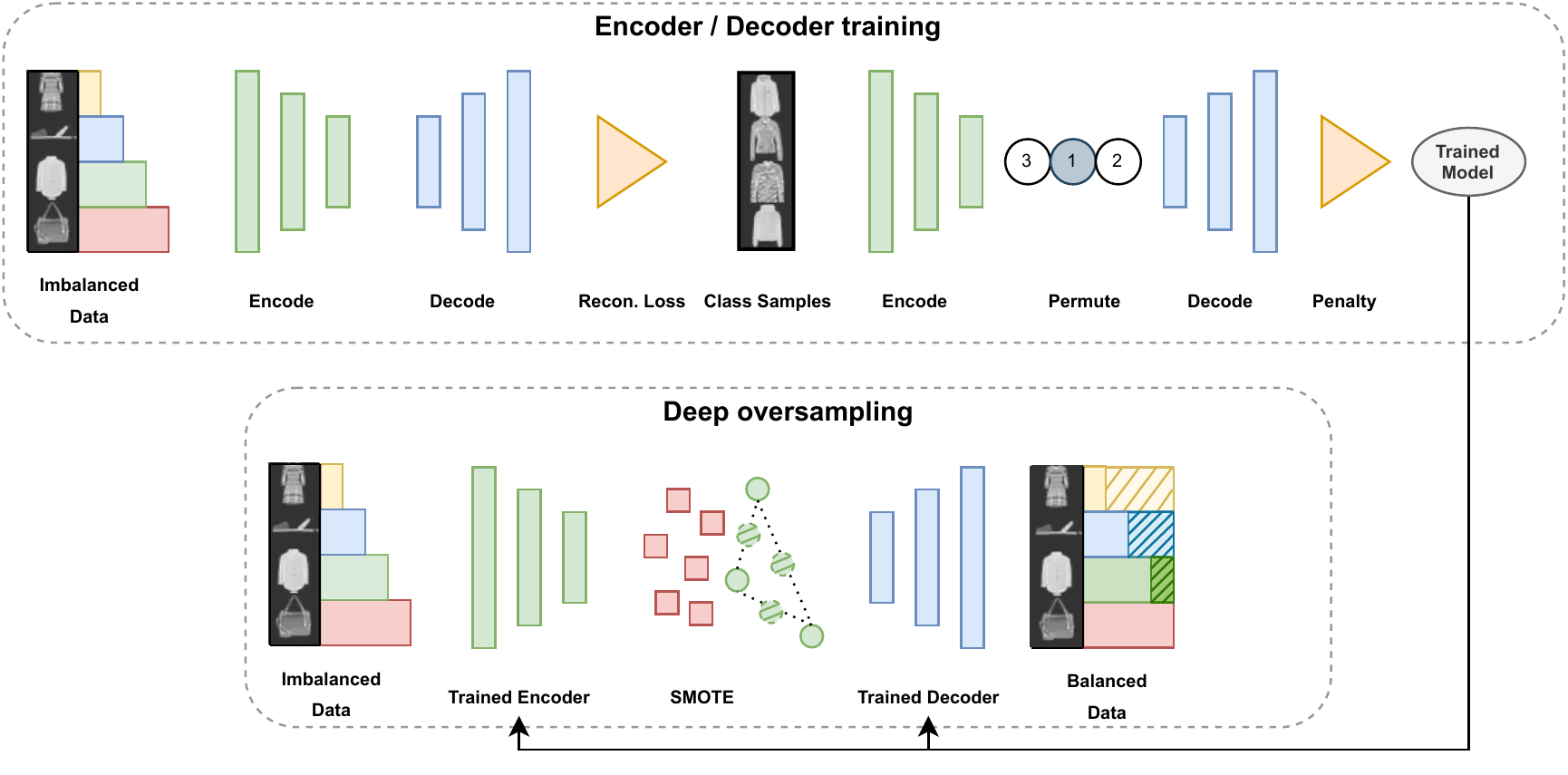}
  \caption{Illustration of DeepSMOTE implementation. The encoder / decoder structure is trained with imbalanced data and a reconstruction and penalty loss. During training, data is sampled, encoded and the order of examples are permuted before decoding.  The trained encoder and decoder are then combined with SMOTE to produce oversampled data.}
  \label{fig:fram}
  \vspace{-0.25cm}
\end{figure}

\begin{figure}[t!]
  \includegraphics[width=0.5\textwidth]{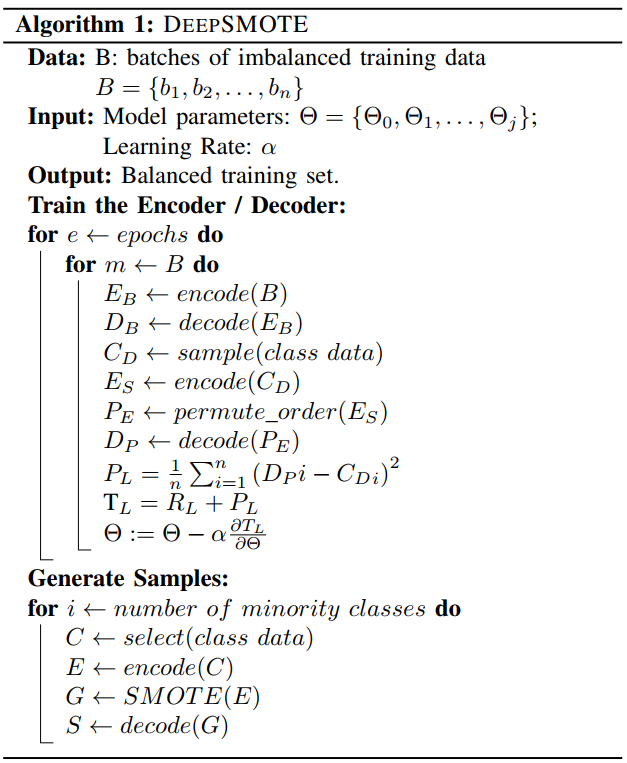}
  \label{fig:fram}
  \vspace{-0.25cm}
\end{figure}

\noindent \textbf{Endoder/decoder framework.} The DeepSMOTE backbone is based on the
 DCGAN architecture, which was established by Radford et al
 \cite{radford2015unsupervised}.  Radford et al. use
 a discriminator / generator in a GAN, which is
 fundamentally similar to an encoder / decoder because the
 discriminator effectively encodes input (absent the final,
 fully connected layer) and the generator (decoder)
 generates output.

The encoder and decoder are trained in
 an end-to-end fashion.  During DeepSMOTE training, an 
imbalanced dataset is fed to the encoder / decoder in
 batches.  A reconstruction loss is computed on the batched
 data.  All classes are used during training so that the
 encoder / decoder can learn to reconstruct both majority
 and minority class images from the imbalanced data.
  Because there are few minority class examples, majority class 
examples are used to train the model to learn the basic
 reconstruction patterns inherent in the data.  This
 approach is based on the assumption that classes
 share some similar characteristics (e.g., all classes
 represent digits or faces).  Thus, for example, although the number
 9 (minority class) resides in a different class than the
 number 0 (majority class), the model learns the basic
 contours of digits.
 
\noindent \textbf{Enhanced loss function.} In addition to a reconstruction loss, the DeepSMOTE loss 
function contains a penalty term.  The penalty term is
 based on a reconstruction of embedded images. DeepSMOTE's penalty loss is produced in the following fashion.
  During training, a batch of images is sampled from the
training set.  The number of sampled images is the same
 as the number of images used for reconstruction loss
 purposes; however, unlike the images used during the
 reconstruction loss phase of training, the sampled images
 are all from the same class.  The sampled images are then
 reduced to a lower dimensional feature space by the
 encoder.  During the decoding phase, the encoded images
 are \emph{not} reconstructed by the decoder in the same
 \emph{order} as the encoded images.  By changing the
 \emph{order} of the reconstructed images, which are all
 from the same class, we effectively introduce \emph{variance}
 into the encoding / decoding process. For example, the encoded order of the images may be D\textsubscript{0},  D\textsubscript{1}, D\textsubscript{2}, and the decoded order of the images may be D\textsubscript{2}, D\textsubscript{0},  D\textsubscript{1}.  This variance
 facilitates the generation of images during inference (where an image is encoded, SMOTEd, and the decoded).
The penalty loss is based on the mean squared error (MSE) difference between D\textsubscript{0} and  D\textsubscript{1}, D\textsubscript{1} and D\textsubscript{2}, etc., as if an image was oversampled by SMOTE (i.e., as if an image were generated based on the difference between an image and the image’s neighbor). This step is designed to insert variance into the encoding / decoding process.
  We therefore obviate the need for a discriminator because we use training data to train the generator
 by simply altering the order
 of the encoded / decoded images.
 
 As a refresher, the SMOTE algorithm generates synthetic instances by randomly selecting a minority class example and one of its class neighbors. The distance between the example and its neighbor is calculated. The distance is multiplied by a random percentage (i.e., between 0 and 1) and added to the example instance in order to generate synthetic instances.  We simulate SMOTE's methodology during DeepSMOTE training by selecting a class sample and calculating a distance between the instance and its neighbors (in the embedding or feature space), except that the distance (MSE) during training is used as an implicit penalty on the reconstruction loss. As noted by Arjovsky et al. \cite{arjovsky2017wasserstein}, many generative deep learning models effectively incorporate a penalty, or noise, term in their loss function, to impart diversity into the model distribution. For example, both VAEs and WAEs include penalty terms in their loss functions.  We use permutation, instead of SMOTE, during training because it is more memory and computationally efficient.  The use of the penalty term, and SMOTE's fidelity in interpolating synthetic samples during the inference phase, allows us to avoid the use of a discriminator, which is typically used by  GAN and WAE models.

\noindent \textbf{Artificial image generation}. Once DeepSMOTE is trained, images can be generated with
 the encoder / decoder structure.  The encoder reduces 
the raw input to a lower dimensional feature space,
 which is oversampled by SMOTE.  The decoder then
 decodes the SMOTEd features into images, which can
 augment the training set of a deep learning classifier.  
 
The main difference between the DeepSMOTE training and generation phases is that during the data generation phase, SMOTE is
 substituted for the order permutation step.  SMOTE is used during data generation to
 introduce variance; whereas,
 during training, variance is introduced by permuting
 the order of the training examples that are encoded
 and then decoded and also through the penalty loss. SMOTE itself does not require training because it is 
non-parametric.    
 
\section{Experimental study}
\label{sec:exp}

We have designed the following experimental study in order to answer the following research questions:

\begin{itemize}
    \item[RQ1:] Is DeepSMOTE capable of outperforming state-of-the-art pixel-based oversampling algorithms?
    \item[RQ2:] Is DeepSMOTE capable of outperforming state-of-the-art GAN-based resampling algorithms designed to work with complex and imbalanced data representations?
    \item[RQ3:] What is the impact of the test set distribution on DeepSMOTE performance? 
    \item[RQ4:] What is the visual quality of artificial images generated by DeepSMOTE? 
    \item[RQ5:] Is DeepSMOTE robust to increasing class imbalance ratios?
    \item[RQ6:] Can DeepSMOTE produce stable models under extreme class imbalance?
\end{itemize}

\subsection{Setup}

\noindent \textbf{Overview of the Datasets.} Five popular datasets were selected as benchmarks for evaluating imbalanced data oversampling:  MNIST \cite{lecun1998gradient}, Fashion-MNIST
 \cite{xiao2017fashion}, CIFAR-10 \cite{krizhevsky2009learning},
 the Street View House Numbers (SVHN) \cite{netzer2011reading},
 and Large-scale CelebFaces Attributes (CelebA)
 \cite{liu2015deep}. Below we discuss their details, while their class distributions are given in Table~\ref{tab:dat}.
 
\noindent \underline{MNIST and Fashion-MNIST}. The MNIST dataset
 consists of handwritten digits and the Fashion-MNIST
 dataset contains Zalando clothing article images.
  Both training sets have 60,000 images and the test
 sets have 10,000 examples.  Both datasets contain
 gray-scale images (1 X 28 X 28), with 10 classes
 each.
  
\noindent \underline{CIFAR-10 and SVHN}. The CIFAR-10 dataset
 consists of images, such as automobiles, cats, dogs,
 frogs and birds, whereas the SVHN dataset consists
 of small, cropped digits from house numbers in Google
 Street View images.  CIFAR-10 has 50,000 training
 images and 10,000 test images. SVHN has   73,257
 digits for training and 26,032 digits for testing.
  Both datasets consist of color images 
(3 X 32 X 32), with 10 classes each.

\noindent \underline{CelebA}.  The CelebA dataset contains 
200,000 celebrity images, each with 40 attribute
 annotations (i.e., classes). The color images
 (3 X 178 X 218) in this dataset cover large pose
 variations and background clutter.  For purposes
 of this study, the images were resized to 
3 X 32 X 32 and 5 classes were selected: black hair,
 brown hair, blond, gray, and bald.
 
     \begin{table}[!h]
 \caption{Class distributions of five benchmark datasets used in experimental evaluation. }
 \centering
\begin{adjustbox}{width=0.48\textwidth,center=0.48\textwidth}
\ra{1.3}
\begin{tabular}{@{}rrrrcrrrcrrr@{}}\toprule
& \multicolumn{3}{c}{$MNIST / FMNIST$} & \phantom{abc}& \multicolumn{3}{c}{$CIFAR / SVHN$} &
\phantom{abc} & \multicolumn{3}{c}{$CELEBA$}\\
\cmidrule{2-4} \cmidrule{6-8} \cmidrule{10-12}
\cmidrule{2-4} \cmidrule{6-8} \cmidrule{10-12}
& Train & Bal. Test & Imbal. Test && Train & Bal. Test & Imbal. Test && Train & Bal. Test & Imbal. Test\\
\midrule
$Class$ & & & & & & & & & & & \\

$0$ & 4000 & 1200 & 1000 && 4500 & 1000 & 1000 && 9000 & 900  & 1000\\
$1$ & 2000 & 1200 & 500  && 2000 & 1000 & 500  && 4500 & 900  & 500\\
$2$ & 1000 & 1200 & 250  && 1000 & 1000 & 250  && 1000 & 900  & 111\\

$3$ & 750  & 1200 & 187  && 800  & 1000 & 187  && 500  & 900  & 55\\
$4$ & 500  & 1200 & 125  && 600  & 1000 & 125  && 160  & 900  & 17\\
$5$ & 350  & 1200 & 87   && 500  & 1000 & 87   &&      &      & \\

$6$ & 200  & 1200 & 50   && 400  & 1000 & 50   &&      &      & \\
$7$ & 100  & 1200 & 25   && 250  & 1000 & 25   &&      &      & \\
$8$ & 60   & 1200 & 15   && 150  & 1000 & 15   &&      &      & \\

$9$ & 40   & 1200 & 10   && 80   & 1000 & 10   &&      &      & \\

\bottomrule
\end{tabular}
\end{adjustbox}
\label{tab:dat}
\end{table}

\begin{figure*}[!h]
  \centering
  \subfloat[Imbalanced data]{\includegraphics[trim={3cm 3cm 3cm 3cm},clip,width=0.24\textwidth]{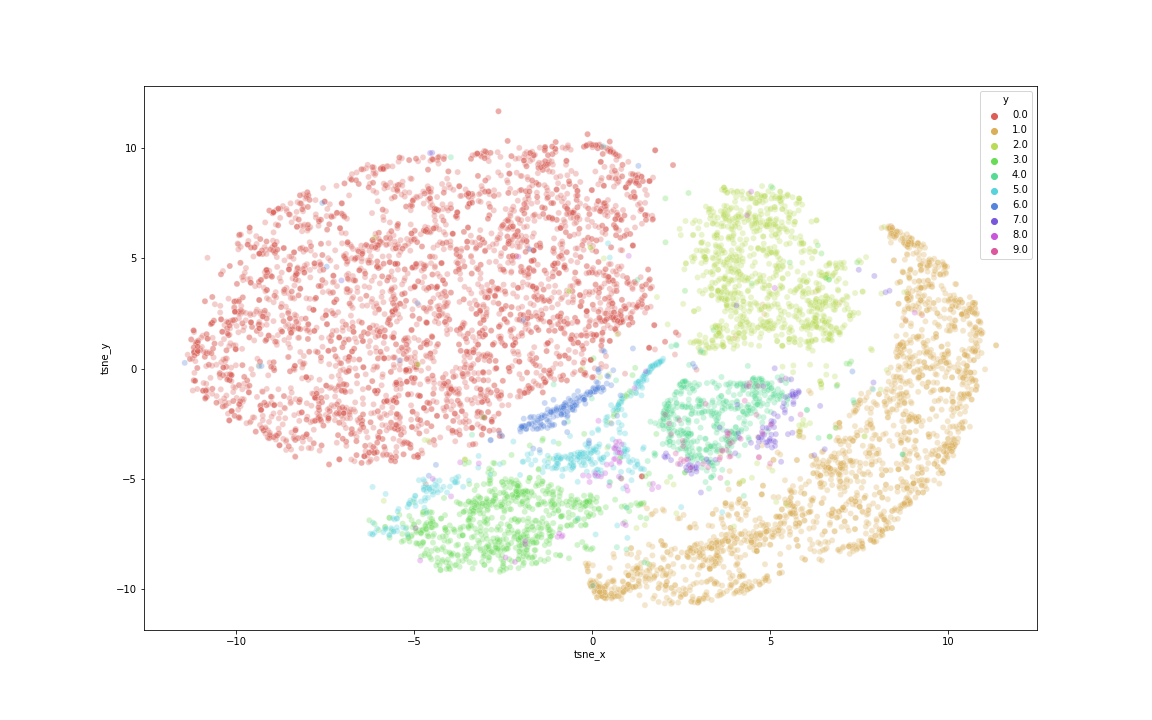}\label{fig:f1}}
  \hfill
  \subfloat[BAGAN]{\includegraphics[trim={3cm 3cm 3cm 3cm},clip,width=0.24\textwidth]{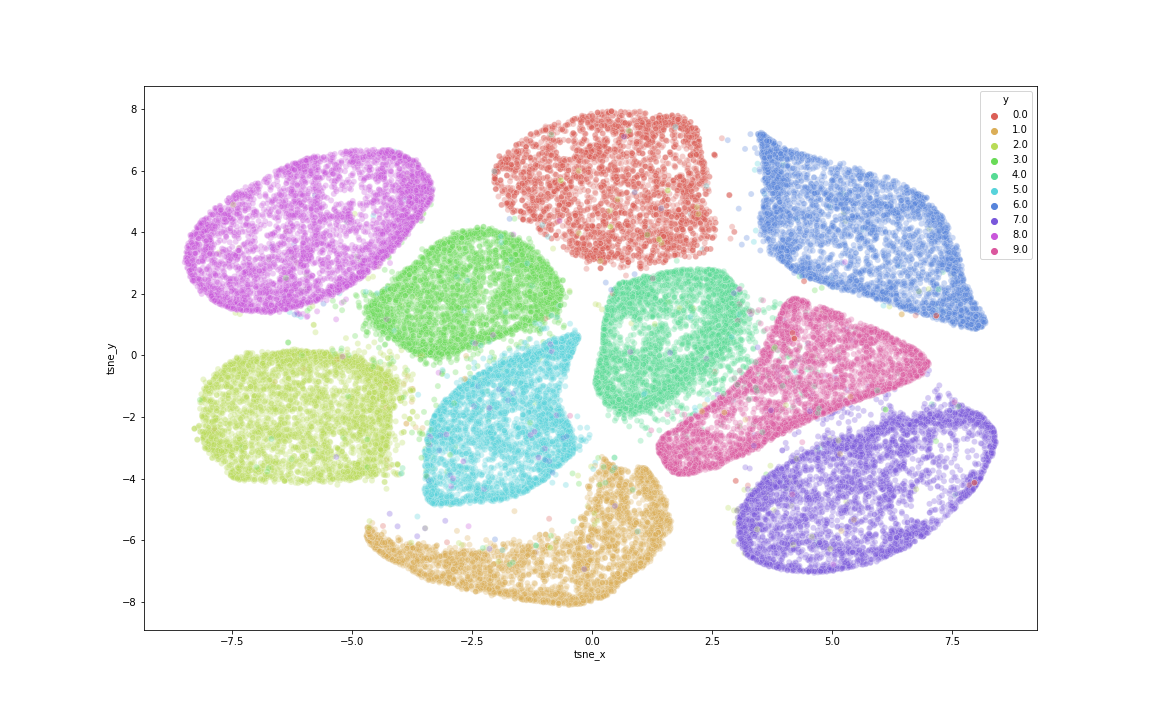}\label{fig:f2}}
   \hfill
  \subfloat[GAMO]{\includegraphics[trim={3cm 3cm 3cm 3cm},clip,width=0.24\textwidth]{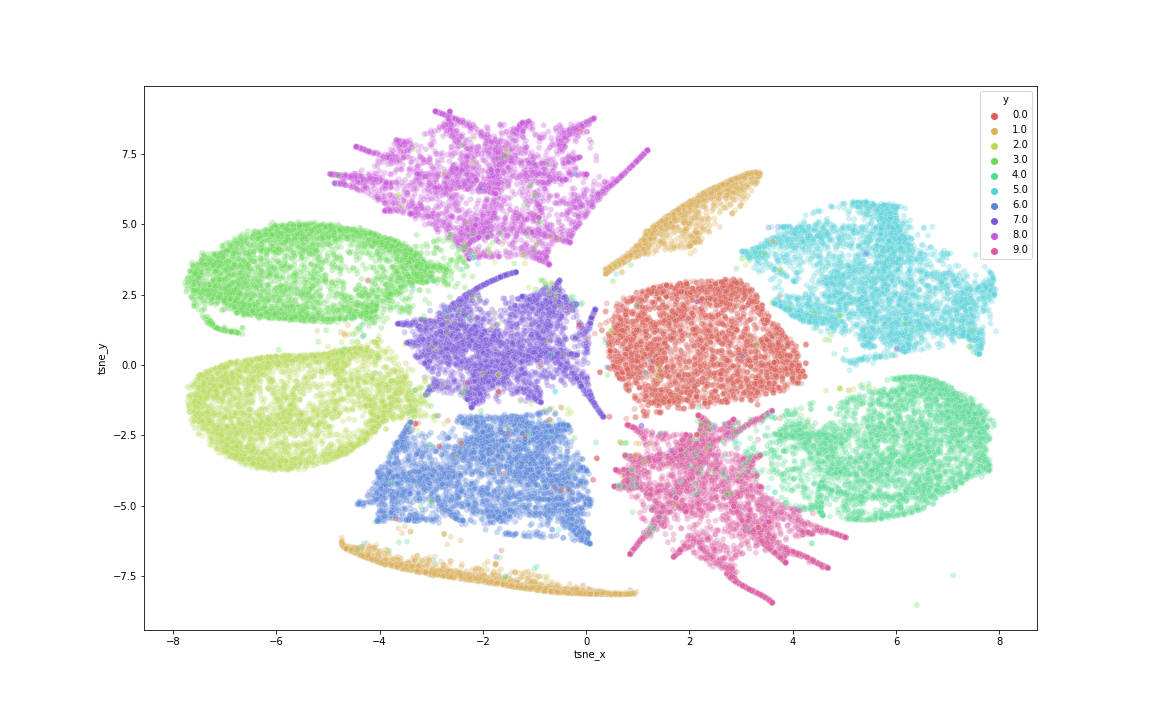}\label{fig:f3}}
   \hfill
  \subfloat[DeepSMOTE]{\includegraphics[trim={3cm 3cm 3cm 3cm},clip,width=0.24\textwidth]{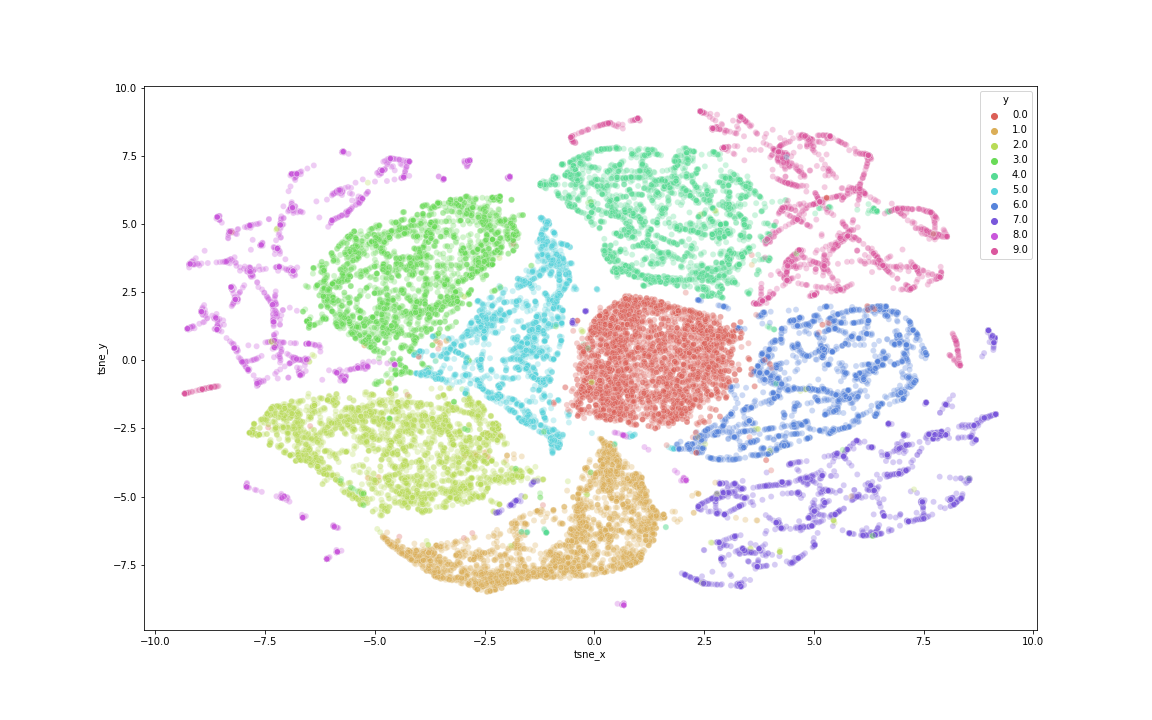}\label{fig:f4}}
  \caption{Illustration of the distribution of the MNIST instances among classes. (a) Original imbalanced training set distribution. (b) Balanced distribution using BAGAN. (c) Balanced distribution with GAMO. (d) Balanced distribution with DeepSMOTE.}
  \label{fig:bal}
\end{figure*}

\smallskip
\noindent \textbf{Introducing class imbalance.} Imbalance was introduced by randomly selecting 
samples from each class in the training sets.  For the MNIST and
 Fashion-MNIST datasets, the number of imbalanced
 examples were: [4000, 2000, 1000, 750, 500, 350,
 200, 100, 60, 40].  For the CIFAR-10 and SVHN 
datasets, the number of imbalanced examples were:
 [4500, 2000, 1000, 800, 600, 500, 400, 250, 150,
 80].   For CelebA, the number of imbalanced
 examples were: [9000, 4500, 1000, 500, 160]. 
For MNIST and Fashion-MNIST, the imbalance ratio
 of the respective majority class compared to the
 smallest minority class was 100:1; and for 
CIFAR-10, SVHN, and CelebA the ratio was approx. 56:1.

\smallskip
\noindent \textbf{Reference resampling methods.} In order to evaluate the effectiveness of DeepSMOTE, we compare it to state-of-the-art shallow and deep resampling methods. We have selected four pixel-based modern oversampling algorithms: SMOTE \cite{chawla2002smote}, Adaptive Mahalanobis Distance-based Oversampling (AMDO) \cite{Yang:2018}, Combined Cleaning and Resampling (MC-CCR) \cite{Koziarski:2020ccr}, and Radial-Based Oversampling (MC-RBO) \cite{Krawczyk:2020}. Additionally, we have chosen two of the top performing GAN-based oversampling approaches:
Balancing GAN (BAGAN) \cite{mariani2018bagan} and Generative
 Adversarial Minority Oversampling (GAMO) \cite{mullick2019generative}. BAGAN initializes its generator with the decoder portion of an autoencoder, which is trained on
 both minority and majority images. GAMO is based on a three-player adversarial game between a convex generator, a classifier network and a discriminator.

\smallskip
\noindent \textbf{Classification model.} All resampling methods use an identical Resnet-18 \cite{he2016deep} as their base classifier.

\smallskip
\noindent \textbf{Performance Metrics.} The following metrics were used to evaluate the
 performance of the various models: Average Class
 Specific Accuracy (ACSA), macro-averaged Geometric
 Mean (GM) and macro-averaged F1 measure (FM).
  Sokolova and Lapalme have demonstrated that
 these measures are not prejudiced toward the
 majority class \cite{sokolova2009systematic}.

\smallskip
\noindent \textbf{Testing procedure.} A 5-fold cross-validation was used for training and testing the evaluated methods. Two approaches to forming test sets were employed: imbalanced and balanced testing. For imbalanced testing, the ratio of test examples follows the same imbalance ratio that exists in the training set (this approach is common in the imbalanced classification domain). With the balanced test sets, the number of test examples was approximately equal across all classes (this approach is common in the long-tailed recognition domain). 

\smallskip
\noindent \textbf{Statistical analysis of results.} In order to assess whether DeepSMOTE returns statistically significantly better results than the reference resampling algorithms, we use the Shaffer post-hoc test \cite{Stapor:2021} and the Bayesian Wilcoxon signed-rank test for statistical comparison over multiple datasets \cite{Benavoli:2017}. Both tests used a statistical significance level of 0.05.

\smallskip
\noindent \textbf{DeepSMOTE Implementation Details.} 
As mentioned above, for DeepSMOTE implementation purposes, we used the DCGAN
architecture developed by Radford et al., with some
modifications.  The encoder structure consists of four convolutional layers, followed by batch normalization \cite{ioffe2015batch}
and the LeakyReLu activation function \cite{maas2013rectifier}.  The hidden
layer dimensions are 64.  The final layer is linear, yielding a latent dimension of 300 for the
MNIST and Fashion-MNIST datasets and 600 for the
CIFAR-10, SVHN and CelebA datasets.
The decoder structure consists of four mirrored  convolutional transpose layers, which use batch normalization and the Rectified Linear Unit (ReLU)
activation function \cite{nair2010rectified},
except for the final layer, which uses Tanh.   We
train the models for 50 to 350 epochs, depending
on when the training loss plateaus.  We use the
Adam optimizer \cite{kingma2014adam}, with a .0002 learning rate. We
implement DeepSMOTE in PyTorch with a NVIDIA GTX-2080 GPU. DeepSMOTE code is publicly available at: \href{https://github.com/dd1github/DeepSMOTE}{https://github.com/dd1github/DeepSMOTE}

  \begin{table*}[h!]
\centering 
\caption{Performance of DeepSMOTE and reference methods on imbalanced test set}
\begin{adjustbox}{width=1\textwidth,center=\textwidth}
\ra{1.3}
\begin{tabular}{@{}rrrrcrrrcrrrcrrrcrrr@{}}\toprule
& \multicolumn{3}{c}{$MNIST$} & \phantom{abc} & \multicolumn{3}{c}{$FMNIST$} & \phantom{abc} & 
\multicolumn{3}{c}{$CIFAR$} & \phantom{abc} &
\multicolumn{3}{c}{$SVHN$} & \phantom{abc} & \multicolumn{3}{c}{$CELEBA$}\\
\cmidrule{2-4} \cmidrule{6-8} \cmidrule{10-12} \cmidrule{14-16} \cmidrule{18-20}
& ACSA & GM & F1 && ACSA & GM & F1 && ACSA & GM & F1 && ACSA & GM & F1 && ACSA & GM & F1  \\
\cmidrule{1-4} \cmidrule{6-8} \cmidrule{10-12} \cmidrule{14-16} \cmidrule{18-20}
SMOTE 		& 81.48 & 83.99 & 82.44 && 67.94 & 74.84 & 67.12 && 28.02 & 50.08 & 29.58 && 70.18 & 76.33 & 71.80 && 60.29 & 70.48 & 60.03 \\
AMDO 		& 84.29 & 88.73 & 84.88 && 74.90 & 80.89 & 75.39 && 31.19 & 53.99 & 32.44 && 71.94 & 78.52 & 73.06 && 63.54 & 72.86 & 62.94 \\
MC-CCR 		& 86.19 & 92.04 & 86.46 && 78.58 & 86.17 & 79.03 && 32.83 & 56.68 & 33.91 && 72.01 & 80.94 & 74.26 && 65.23 & 77.14 & 64.88 \\
MC-RBO 		& 87.25 & 94.46 & 88.69 && 80.06 & 88.02 & 80.14 && 33.01 & 59.15 & 35.83 && 74.20 & 82.97 & 74.91 && 67.11 & 80.52 & 65.37\\
\cmidrule{1-4} \cmidrule{6-8} \cmidrule{10-12} \cmidrule{14-16} \cmidrule{18-20}
BAGAN 		& 92.56 & 96.11 & 93.85 && 82.50 & 90.51 & 82.96 && 42.41 & 64.12 & 43.01 && 75.81 & 86.44 & 77.02 && 68.62 & 80.84 & \textbf{68.33} \\
GAMO 		& 95.45 & 97.61 & 95.11 && 83.05 & 90.76 & 83.00 && 44.72 & 65.72 & \textbf{45.93} && 75.07 & 86.00 & 76.68 && 66.06 & 79.11 & 64.85 \\
\cmidrule{1-4} \cmidrule{6-8} \cmidrule{10-12} \cmidrule{14-16} \cmidrule{18-20}
DeepSMOTE 	& \textbf{96.16} & \textbf{98.11} & \textbf{96.44} && \textbf{84.88} & \textbf{91.63} & \textbf{83.79} && \textbf{45.26} & \textbf{66.13} & 44.86 && \textbf{79.59} & \textbf{88.67} & \textbf{80.71} && \textbf{72.40} & \textbf{82.91} & 66.99 \\
\bottomrule
\end{tabular}
\end{adjustbox}
\label{tab:imb}
\end{table*}

 \begin{table*}[h!]
\centering 
\caption{Performance of DeepSMOTE and  reference methods on balanced test set (long-tailed recognition setup}
\begin{adjustbox}{width=1\textwidth,center=\textwidth}
\ra{1.3}
\begin{tabular}{@{}rrrrcrrrcrrrcrrrcrrr@{}}\toprule
& \multicolumn{3}{c}{$MNIST$} & \phantom{abc} & \multicolumn{3}{c}{$FMNIST$} & \phantom{abc} & 
\multicolumn{3}{c}{$CIFAR$} & \phantom{abc} &
\multicolumn{3}{c}{$SVHN$} & \phantom{abc} & \multicolumn{3}{c}{$CELEBA$}\\
\cmidrule{2-4} \cmidrule{6-8} \cmidrule{10-12} \cmidrule{14-16} \cmidrule{18-20}
& ACSA & GM & F1 && ACSA & GM & F1 && ACSA & GM & F1 && ACSA & GM & F1 && ACSA & GM & F1  \\
\cmidrule{1-4} \cmidrule{6-8} \cmidrule{10-12} \cmidrule{14-16} \cmidrule{18-20}
SMOTE 		& 87.98 & 89.99 & 85.02 && 70.58 & 76.39 & 68.06 && 27.93 & 42.81 & 25.10 && 68.19 & 74.48 & 64.28 && 48.19 & 56.39 & 42.19\\
AMDO 		& 88.34 & 91.03 & 87.28 && 72.98 & 79.36 & 71.53 && 31.85 & 48.19 & 30.04 && 71.59 & 79.13 & 68.47 && 51.44 & 60.73 & 47.28\\
MC-CCR 		& 90.83 & 93.18 & 91.22 && 75.78 & 81.04 & 74.39 && 33.48 & 51.18 & 32.88 && 74.29 & 81.62 & 72.49 && 58.46 & 65.39 & 57.91\\
MC-RBO 		& 91.28 & 94.62 & 92.49 && 76.91 & 82.14 & 75.92 && 39.17 & 59.29 & 40.37 && 75.38 & 81.98 & 73.52 && 61.53 & 72.95 & 62.08\\
\cmidrule{1-4} \cmidrule{6-8} \cmidrule{10-12} \cmidrule{14-16} \cmidrule{18-20}
BAGAN 		& 93.06 & 95.98 & 92.77	&& 81.48 & 89.31 & 80.93 && 43.38 & 63.73 & 40.25 && 80.23 & 86.77 & 77.75 && 66.09 & 77.77 & 62.84\\
GAMO 		& 95.52 & 97.47 & 95.47 && 83.03 & 90.26 & 82.50 && 44.89 & 65.30 & 43.35 && 80.53 & 87.17 & 78.21 && 66.00 & 77.71 & 63.01\\
\cmidrule{1-4} \cmidrule{6-8} \cmidrule{10-12} \cmidrule{14-16} \cmidrule{18-20}
DeepSMOTE 	& \textbf{96.09} & \textbf{97.80} & \textbf{96.03} && \textbf{83.63} & \textbf{90.61} & \textbf{83.27} && \textbf{45.38} & \textbf{65.30} & \textbf{43.35} && \textbf{80.94} & \textbf{87.39} & \textbf{78.73} && \textbf{69.88} & \textbf{80.38} & \textbf{69.19}\\
\bottomrule
\end{tabular}
\end{adjustbox}
\label{tab:bal}
\end{table*} 

\begin{table}[h!]
	\caption{Results of Shaffer post-hoc tests and Bayesian Wilcoxon signed-rank tests with respect to $p$-values for pairwise comparison between DeepSMOTE and the reference oversampling based methods for three performance metrics. We merged results from imbalanced and long-tailed recognition test scenarios.}
	\label{tab:shf}
	\centering
		\setlength{\tabcolsep}{3pt}
		\scalebox{0.91}{
		\begin{tabular}{l  c  c  c c  c c c}
			\toprule
			DeepSMOTE& \multicolumn{3}{c}{Shaffer post-hoc} && \multicolumn{3}{c}{Bayesian Wilcoxon signed-rank}\\
			 vs. &ACSA &GM &F1 & \phantom{abc} &ACSA &GM &F1\\ 
		\cmidrule(lr){1-1} \cmidrule(lr){2-4} \cmidrule{6-8}
		 SMOTE  &0.00001 & 0.00000 & 0.00001 & &0.00001 & 0.00000 & 0.00001\\
		AMDO & 0.00316 & 0.00048 & 0.00329 & &0.00172 & 0.00026 & 0.00188 \\
		MC-CCR  &0.01042 & 0.00072 & 0.01003 & &0.00099 & 0.00018 & 0.00083 \\
		MC-RBO & 0.02141& 0.01625 &0.02331 & &0.02007 & 0.01002 & 0.02106\\
		\cmidrule(lr){1-1} \cmidrule(lr){2-4} \cmidrule{6-8}
		BAGAN & 0.03148& 0.01352 & 0.03319 & &0.02581 & 0.01039 & 0.02606 \\
	    GAMO& 0.03204 & 0.01488 & 0.03582 & &0.02620 & 0.01721 & 0.02938\\
			\bottomrule
		\end{tabular}
		}
\end{table}

\subsection{Experiment 1: Comparison with state-of-the-art}

\smallskip
\noindent \textbf{Placement of artificial instances.} One of the crucial elements of oversampling algorithms based on artificial instance generation lies in where in the feature space they place their instances. Random positioning is far from desirable, as we want to maintain the original properties of minority classes and enhance them in uncertain / difficult regions. Those regions are mostly class borders, overlapping areas, and small disjuncts. Therefore, the best oversampling methods focus on smart placement of instances that not only balances class distributions, but also reduces the learning difficulty. Figure~\ref{fig:bal} depicts a 2D projection of an imbalanced MNIST dataset, as well as the class distributions after oversampling with BAGAN, GAMO, and DeepSMOTE. We can notice that both BAGAN and GAMO concentrate on saturating the distribution of each class independently, generating a significant number of artificial instances within the main distribution of each class. Such an approach balances the training data and may be helpful for some density-based classifiers. However, neither BAGAN nor GAMO focus on introducing artificial instances in a directed fashion to enhance class boundaries and improve the discrimination capabilities of a classifier trained on oversampled data. DeepSMOTE combines oversampling controlled by the class geometry with our penalty function to introduce instances in such a way that the error probability is reduced on minority classes. This leads to better placement of artificial instances and, as seen in the experimental comparison, more accurate classification.  

\smallskip
\noindent \textbf{Comparison with pixel-based oversampling.} The first group of reference algorithms are four state-of-the-art oversampling approaches. Tables~\ref{tab:imb} and~\ref{tab:bal} show their results for three metrics and two test set distribution types. We can clearly see that pixel-based oversampling is inferior to both GAN-based algorithms and DeepSMOTE. This allows us to conclude that pixel-based oversampling is not a good choice when dealing with complex and imbalanced images. Unsurprisingly, standard SMOTE performs worst of all of the evaluated algorithms, while three other methods try to offset their inability to handle spatial properties of data with advanced instance generation modules. Both MC-CCR and MC-RBO return the best results from all four tested algorithms, with MC-RBO coming close to GAN-based methods. This can be attributed to their compound oversampling solutions, which analyze the difficulty of instances and optimize the placement of new instances, while cleaning overlapping areas. However, this comes at the cost of very high computational complexity and challenging parameter tuning. DeepSMOTE returns superior balanced training sets compared to pixel-based approaches, while providing an intuitive and easy to tune architecture and, according to both non-parametric and Bayesian tests presented in Table~\ref{tab:shf}, outperforms all pixel-based approaches in a statistically significant manner (\textbf{RQ1 answered}).

\smallskip
\noindent \textbf{Comparison with GAN-based oversampling.} Tables~\ref{tab:imb} and~\ref{tab:bal} show that regardless of the metric used, DeepSMOTE outperforms the baseline GAN-based models on all but two cases. Both of these situations are happening with F1 measure and for different models (BAGAN displays a slightly higher F1 value on CelebA, while GAMO on CIFAR). It is important to note that for the same benchmarks, DeepSMOTE offers significantly higher ACSA and GM values than any of these reference algorithms, allowing us to conclude that F1 performance variation is not reflective on how DeepSMOTE can handle minority classes. The success of DeepSMOTE can be attributed to better placement of artificial instances and empowering uncertainty areas because oversampling is driven by our penalized loss function. DeepSMOTE can enhance decision boundaries, effectively reducing the classifier bias towards the majority classes. Furthermore, DeepSMOTE does not share some of the limitations of GAN-based oversampling, such as mode collapse. As DeepSMOTE is driven by the SMOTE-based approach for selecting and placing artificial instances, we ensure that the minority classes are enriched with diverse training data of high discriminative quality. Table~\ref{tab:shf} shows that DeepSMOTE outperforms all GAN-based approaches in a statistically significant manner (\textbf{RQ2 answered}). This comes with an additional gain of directly generating higher-quality artificial images (as will be discussed in the following experiment).

\noindent We note that the CIFAR-10 dataset was the most challenging benchmark for deep oversampling algorithms. We hypothesize that the reason why the models did not exhibit high accuracy on CIFAR-10 compared to the other datasets is because the CIFAR-10 classes do not have similar attributes.  For example, in MNIST and SVHN, all classes are instances of digits and in the case of CelebA, all classes represent faces; whereas, in CIFAR-10, the classes are diverse (e.g., cat, dog, airplane, frog).  Therefore, the models are not able to leverage information that they learn from the majority class (which has more examples) to the minority class (which contains fewer examples). In addition, we also noticed that, in some cases, there appears to be a significant overlap of CIFAR-10 class features. 

\smallskip
\noindent \textbf{Effects of test set distribution.} The final part of the first experiment focused on evaluating the role of class distributions in the test set. In the domain of learning from imbalanced data, the test set follows the distribution of the training set, in order to reflect the actual class disproportions \cite{krawczyk2016learning}. This also impacts the calculation of several cost-sensitive measures that more severely penalize the errors on minority classes \cite{Fernandez:2018}. However, the recently emerging field of long-tailed recognition follows a different testing protocol \cite{Tan:2020}. In this scenario of extreme multi-class imbalance, the training set is skewed, but test sets for most benchmarks are balanced. As DeepSMOTE aims to be a universal approach for imbalanced data preprocessing and resampling, we evaluated its performance in both scenarios. Table~\ref{tab:imb} reports results for the traditional imbalanced setup, while Table ~\ref{tab:bal} reflects the long-tailed recognition setup.  We can see that DeepSMOTE excels in both scenarios, confirming our previous observations on its benefits over pixel-based and GAN-based approaches. It is interesting to see that for the long-tailed setup, DeepSMOTE returns slightly better F1 performance on the CIFAR10 and CelebA datasets. This can be explained by the way the F1 measure is calculated, as it gives equal importance to precision and recall. When dealing with a balanced test set, DeepSMOTE was able to return even better performance on these two metrics. For all other metrics and datasets, DeepSMOTE showcases similar trends for imbalanced and balanced test sets. This allows us to conclude that DeepSMOTE is a suitable and effective solution for both imbalanced and long-tailed recognition scenarios (\textbf{RQ3 answered}).

\subsection{Experiment 2: Quality of artificially generated images}

\begin{figure*}[!h]
  \centering
  \subfloat[Originals]{\includegraphics[width=0.2\textwidth]{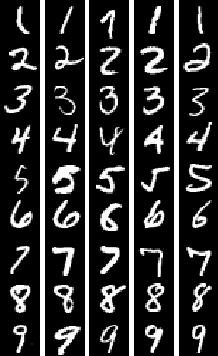}\label{fig:f1}}
  \hfill
  \subfloat[BAGAN]{\includegraphics[width=0.2\textwidth]{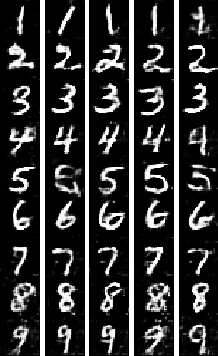}\label{fig:f2}}
   \hfill
  \subfloat[GAMO]{\includegraphics[width=0.2\textwidth]{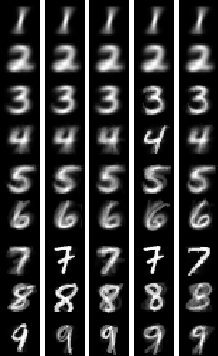}\label{fig:f2}}
   \hfill
  \subfloat[DeepSMOTE]{\includegraphics[width=0.2\textwidth]{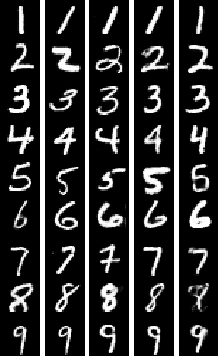}\label{fig:f2}}
  \caption{MNIST minority class images, with rows corresponding to digit classes}
  \label{fig:gen1}
\end{figure*}

\begin{figure*}[!h]
  \centering
  \subfloat[Originals]{\includegraphics[width=0.2\textwidth]{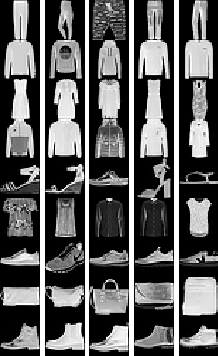}\label{fig:f1}}
  \hfill
  \subfloat[BAGAN]{\includegraphics[width=0.2\textwidth]{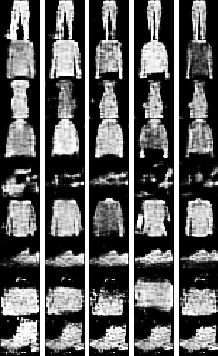}\label{fig:f2}}
   \hfill
  \subfloat[GAMO]{\includegraphics[width=0.2\textwidth]{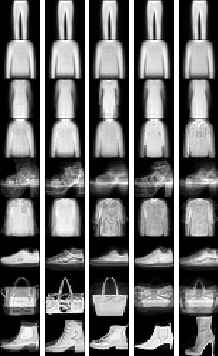}\label{fig:f2}}
   \hfill
  \subfloat[DeepSMOTE]{\includegraphics[width=0.2\textwidth]{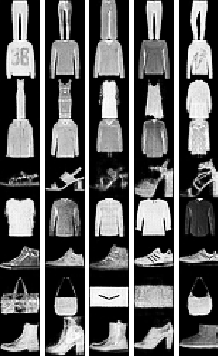}\label{fig:f2}}
  \caption{Fashion MNIST minority class images: trouser / pullover / dress / coat / sandal / shirt / sneaker / bag / ankle boot}
\end{figure*}

\begin{figure*}[!h]
  \centering
  \subfloat[Originals]{\includegraphics[width=0.2\textwidth]{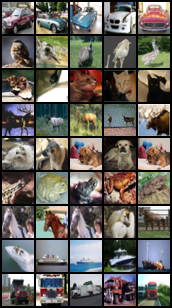}\label{fig:f1}}
  \hfill
  \subfloat[BAGAN]{\includegraphics[width=0.2\textwidth]{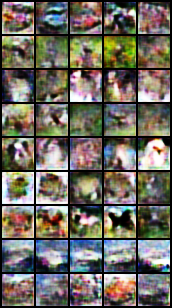}\label{fig:f2}}
   \hfill
  \subfloat[GAMO]{\includegraphics[width=0.2\textwidth]{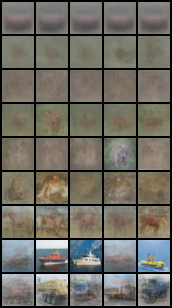}\label{fig:f2}}
   \hfill
  \subfloat[DeepSMOTE]{\includegraphics[width=0.2\textwidth]{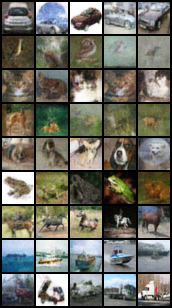}\label{fig:f2}}
  \caption{CIFAR-10 minority class images: automobile / bird / cat / deer / dog / frog / horse / ship / truck}
\end{figure*}

\begin{figure*}[!h]
  \centering
  \subfloat[Originals]{\includegraphics[width=0.2\textwidth]{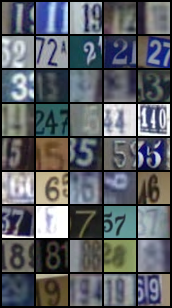}\label{fig:f1}}
  \hfill
  \subfloat[BAGAN]{\includegraphics[width=0.2\textwidth]{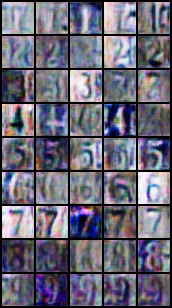}\label{fig:f2}}
   \hfill
  \subfloat[GAMO]{\includegraphics[width=0.2\textwidth]{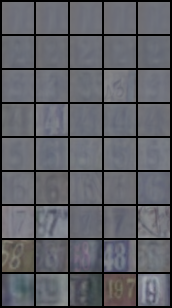}\label{fig:f2}}
   \hfill
  \subfloat[DeepSMOTE]{\includegraphics[width=0.2\textwidth]{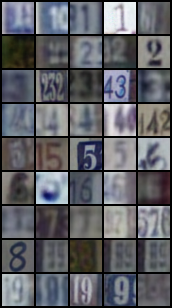}\label{fig:f2}}
  \caption{SVHN minority class images,with rows corresponding to digit classes}
\end{figure*}

\begin{figure*}[!h]
  \centering
  \subfloat[Originals]{\includegraphics[width=0.2\textwidth]{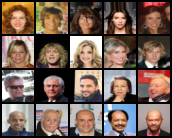}\label{fig:f1}}
  \hfill
  \subfloat[BAGAN]{\includegraphics[width=0.2\textwidth]{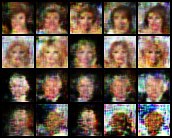}\label{fig:f2}}
   \hfill
  \subfloat[GAMO]{\includegraphics[width=0.2\textwidth]{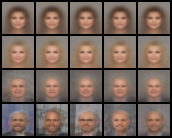}\label{fig:f2}}
   \hfill
  \subfloat[DeepSMOTE]{\includegraphics[width=0.2\textwidth]{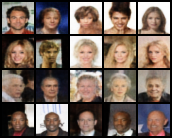}\label{fig:f2}}
  \caption{CELEBA minority class images: brown hair / blond hair / gray hair / bald}
  \label{fig:gen5}
\end{figure*}

Figures~\ref{fig:gen1} to~\ref{fig:gen5} presents the artificially generated images for all five benchmark datasets by BAGAN, GAMO, and the DeepSMOTE. We can see the quality of DeepSMOTE-generated images. This can be attributed to DeepSMOTE using an efficient encoding/decoding architecture with an enhanced loss function, as well as preserving class topology via metric-based instance imputation.  We note that in the case of GAMO, we present images that were used for classification purposes and not images generated by the GAMO2PIX method, so as to provide a direct comparison of GAMO training images to training images generated by BAGAN and DeepSMOTE. The outcomes of both experiments demonstrate that DeepSMOTE generates artificial images that are both information-rich (i.e., they improve the  discriminative ability of deep classifiers and they counter majority bias) and are of high visual quality (\textbf{RQ4 answered}).

\subsection{Experiment 3: Robustness and stability under varied imbalance ratios}

\begin{figure*}[!tbp]
  \includegraphics[width=0.99\textwidth]{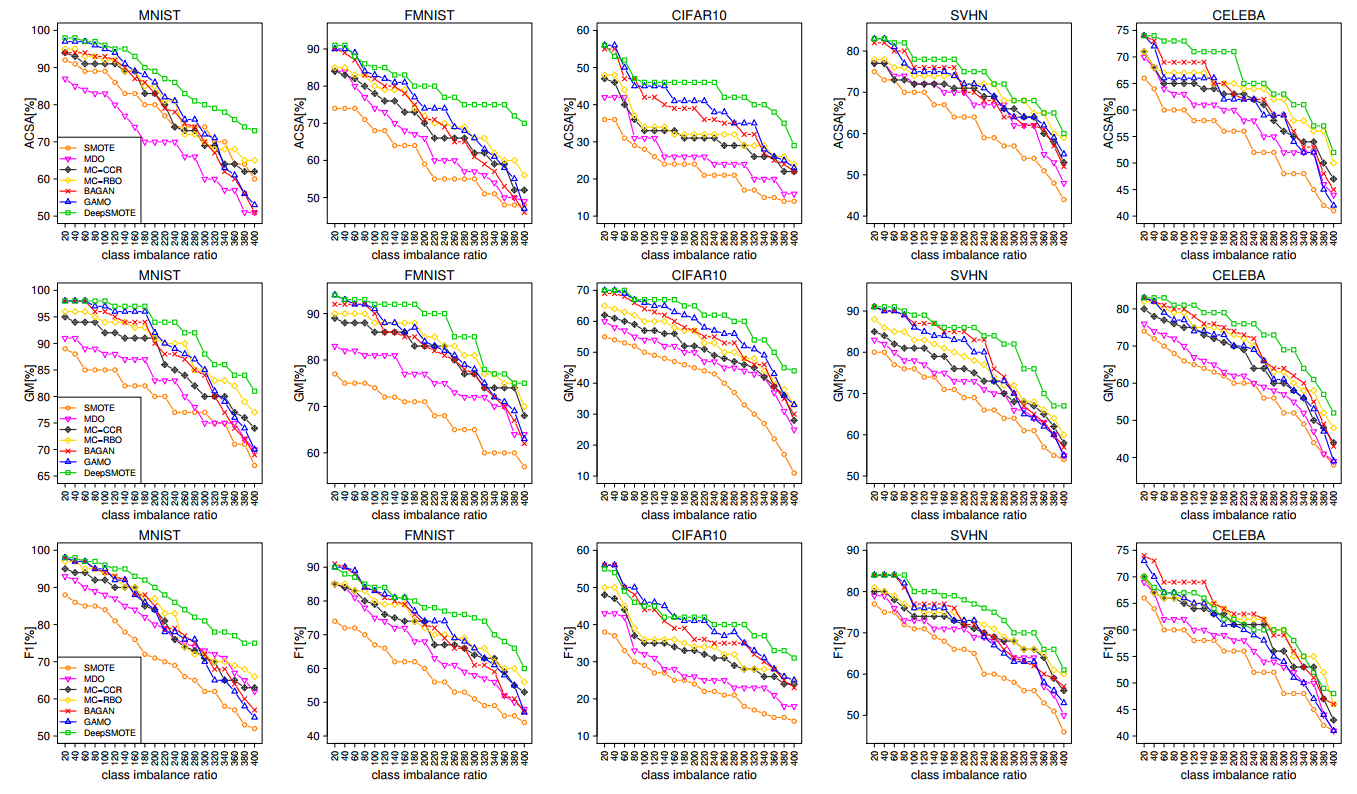}
  \caption{Robustness to increasing imbalance ratios for DeepSMOTE and reference resampling methods}
  \label{fig:rob}
\end{figure*}

\begin{figure*}[!tbp]
  \includegraphics[width=0.19\textwidth]{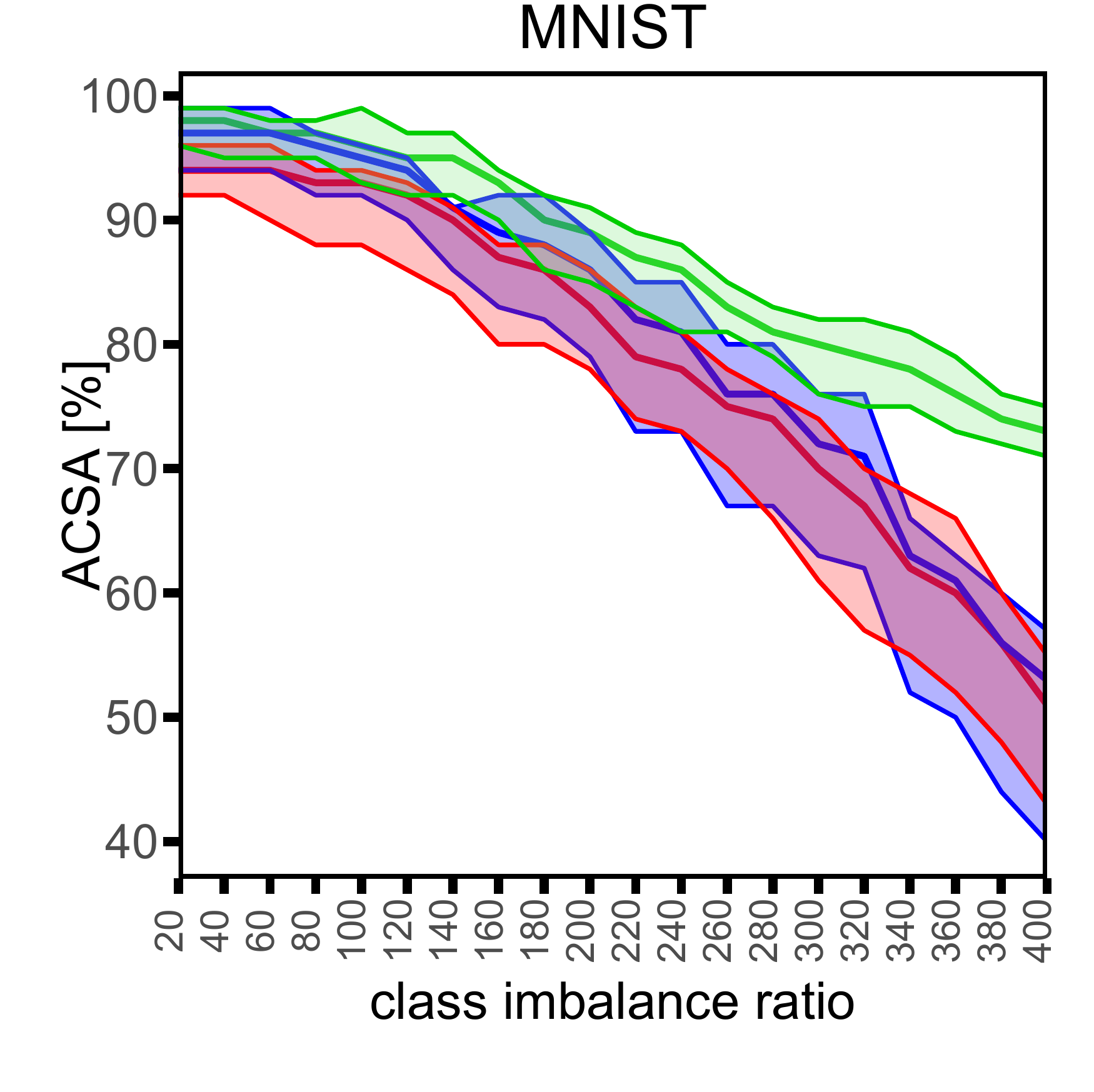}
  \includegraphics[width=0.19\textwidth]{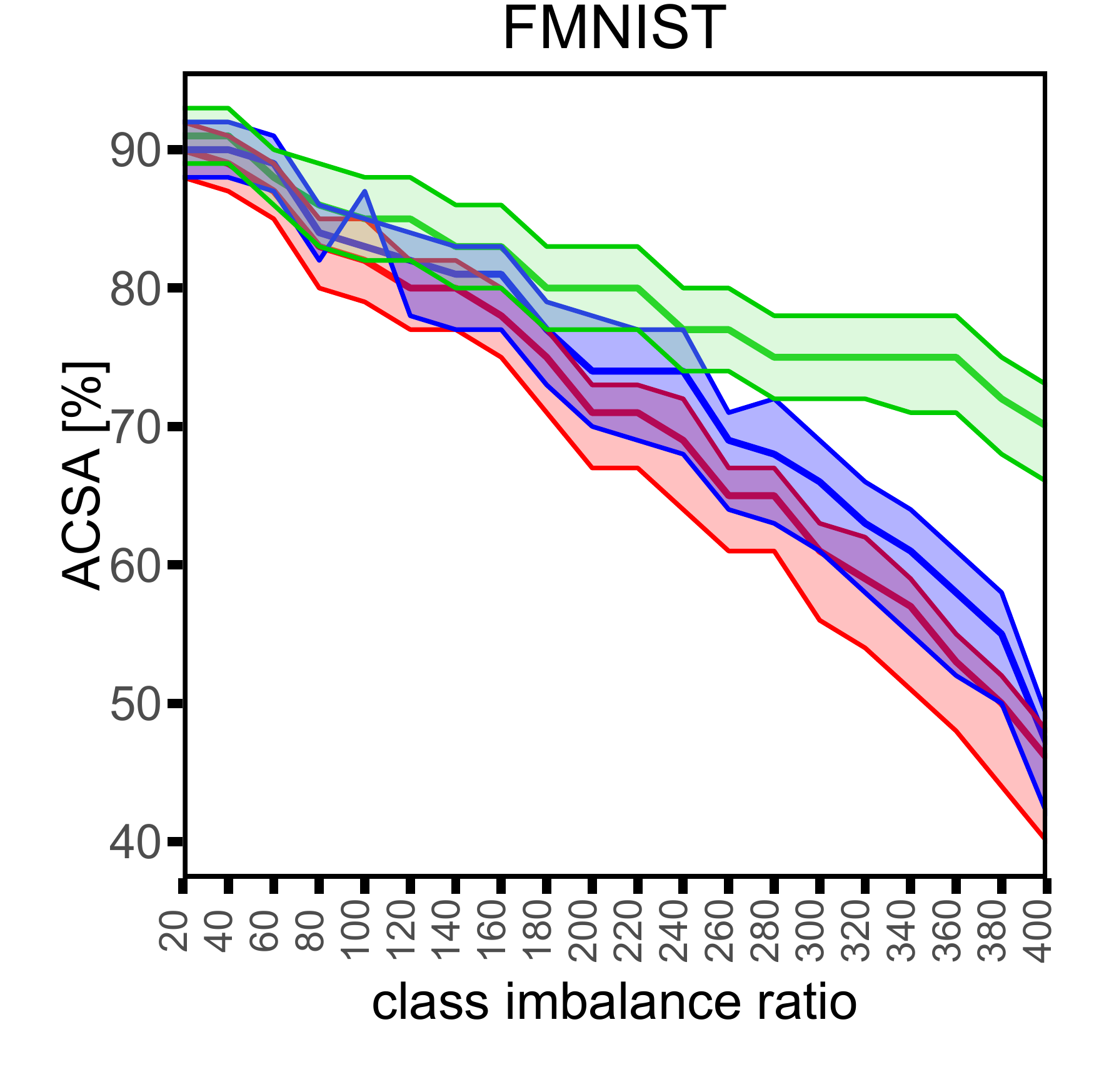}
  \includegraphics[width=0.19\textwidth]{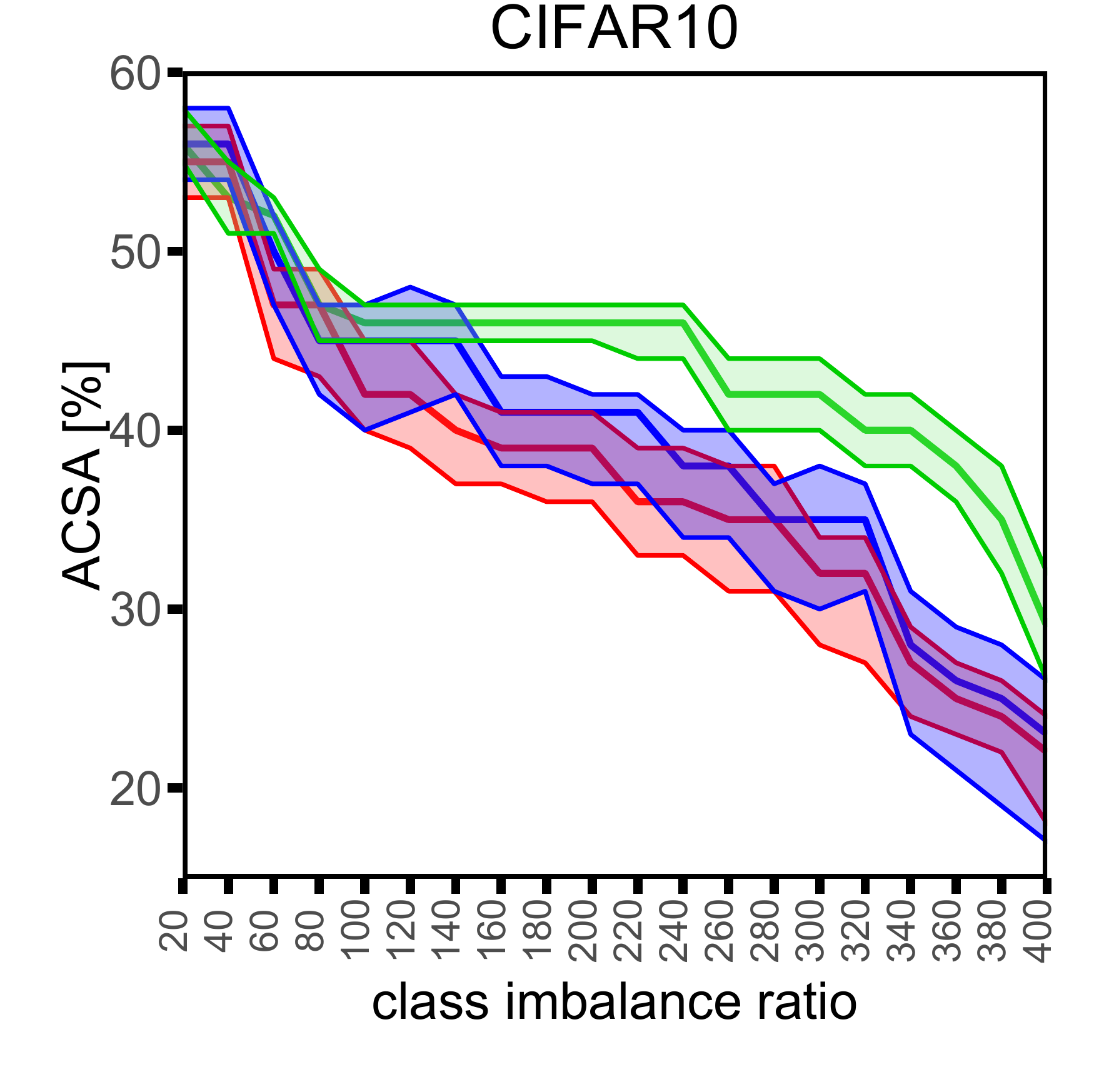}
  \includegraphics[width=0.19\textwidth]{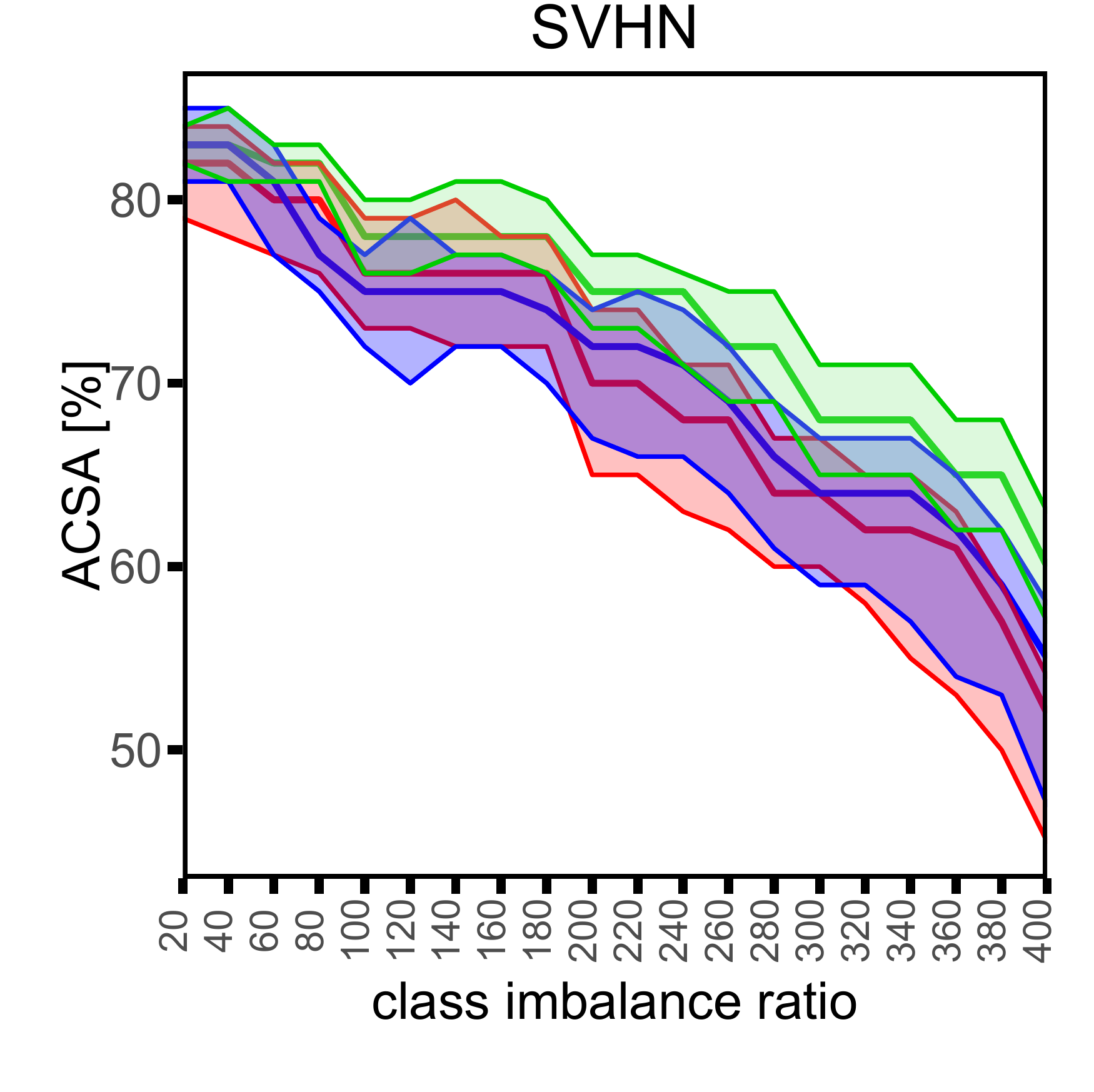}
  \includegraphics[width=0.19\textwidth]{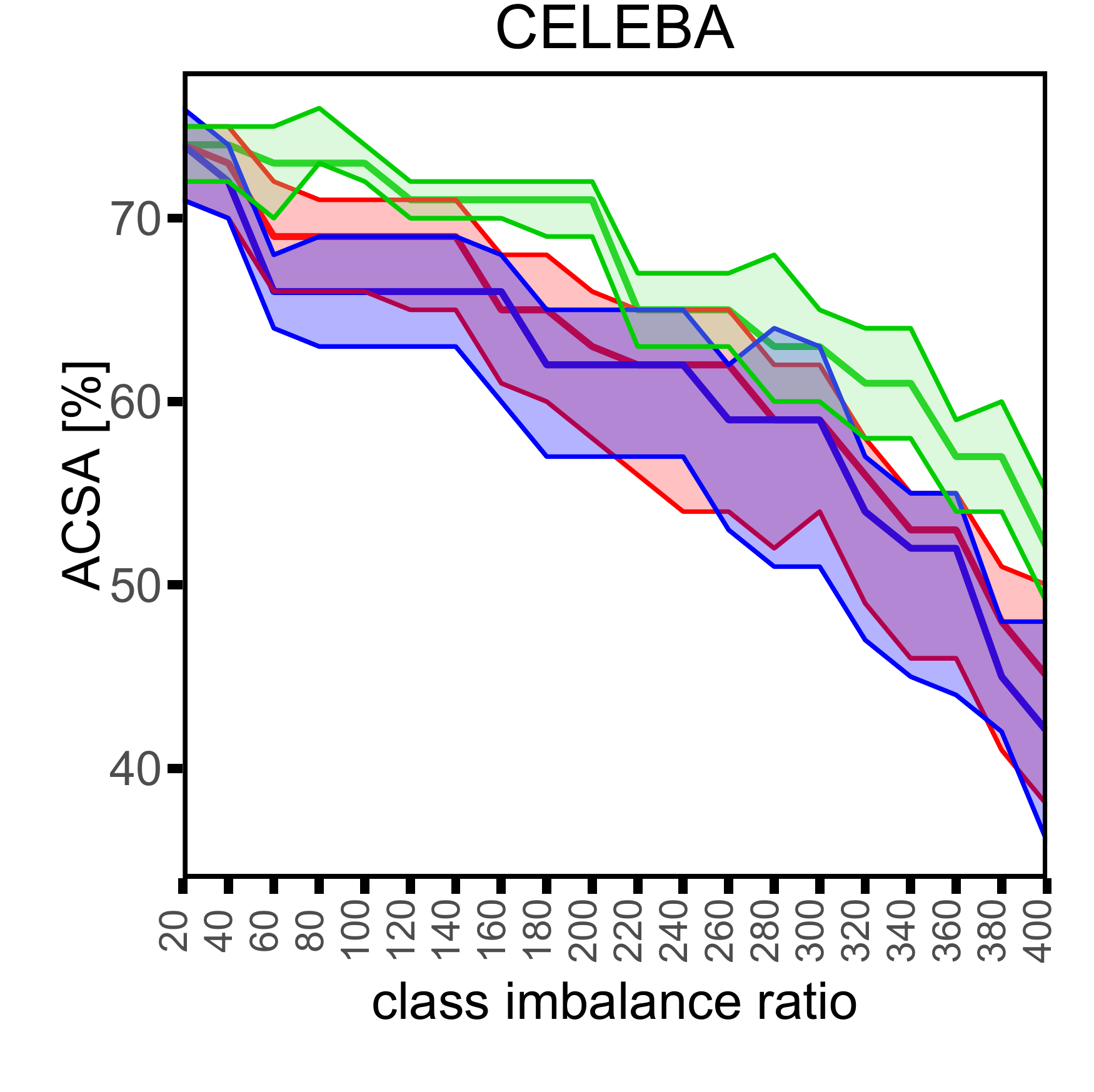}
 
  \includegraphics[width=0.19\textwidth]{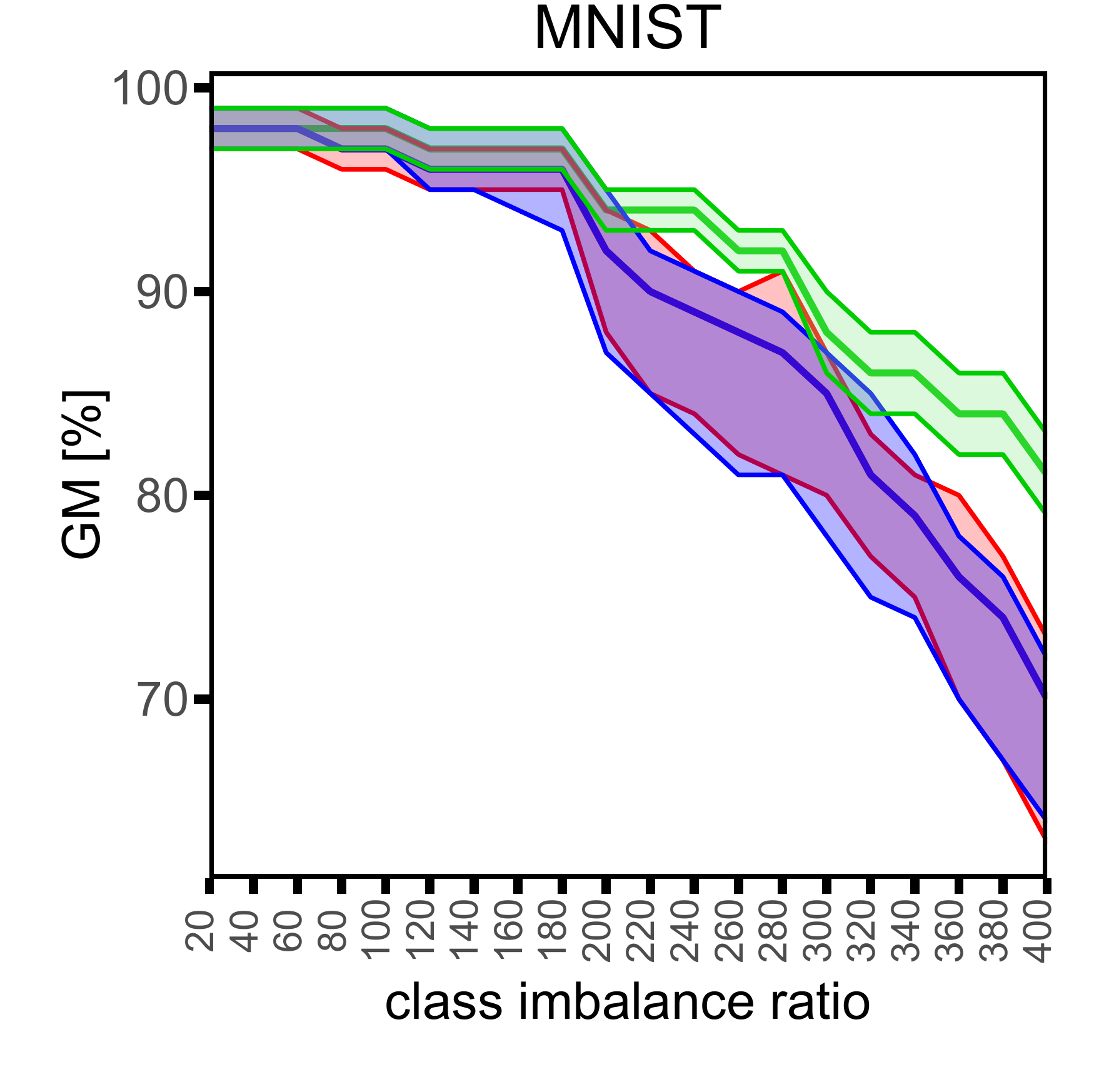}
  \includegraphics[width=0.19\textwidth]{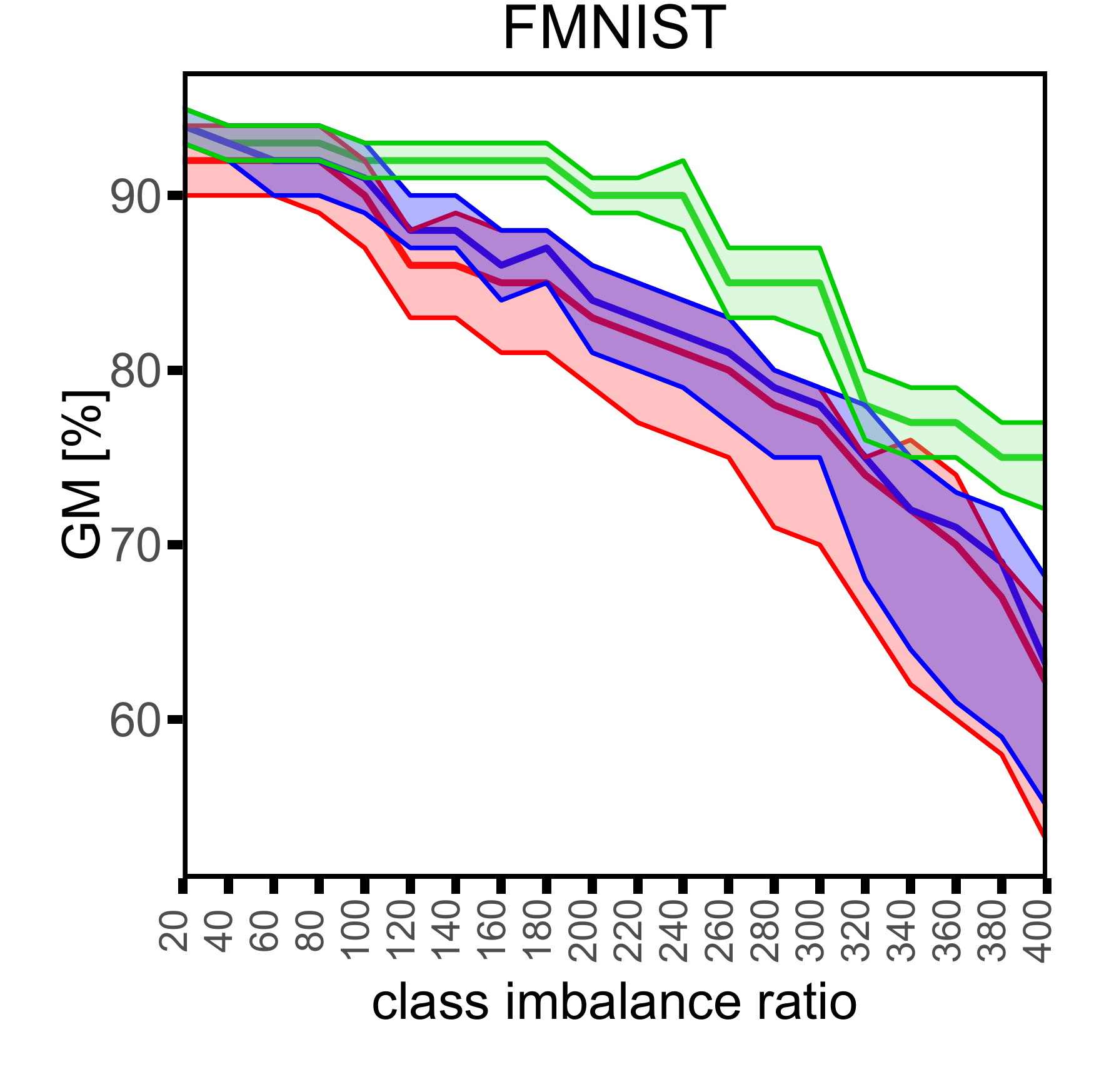}
  \includegraphics[width=0.19\textwidth]{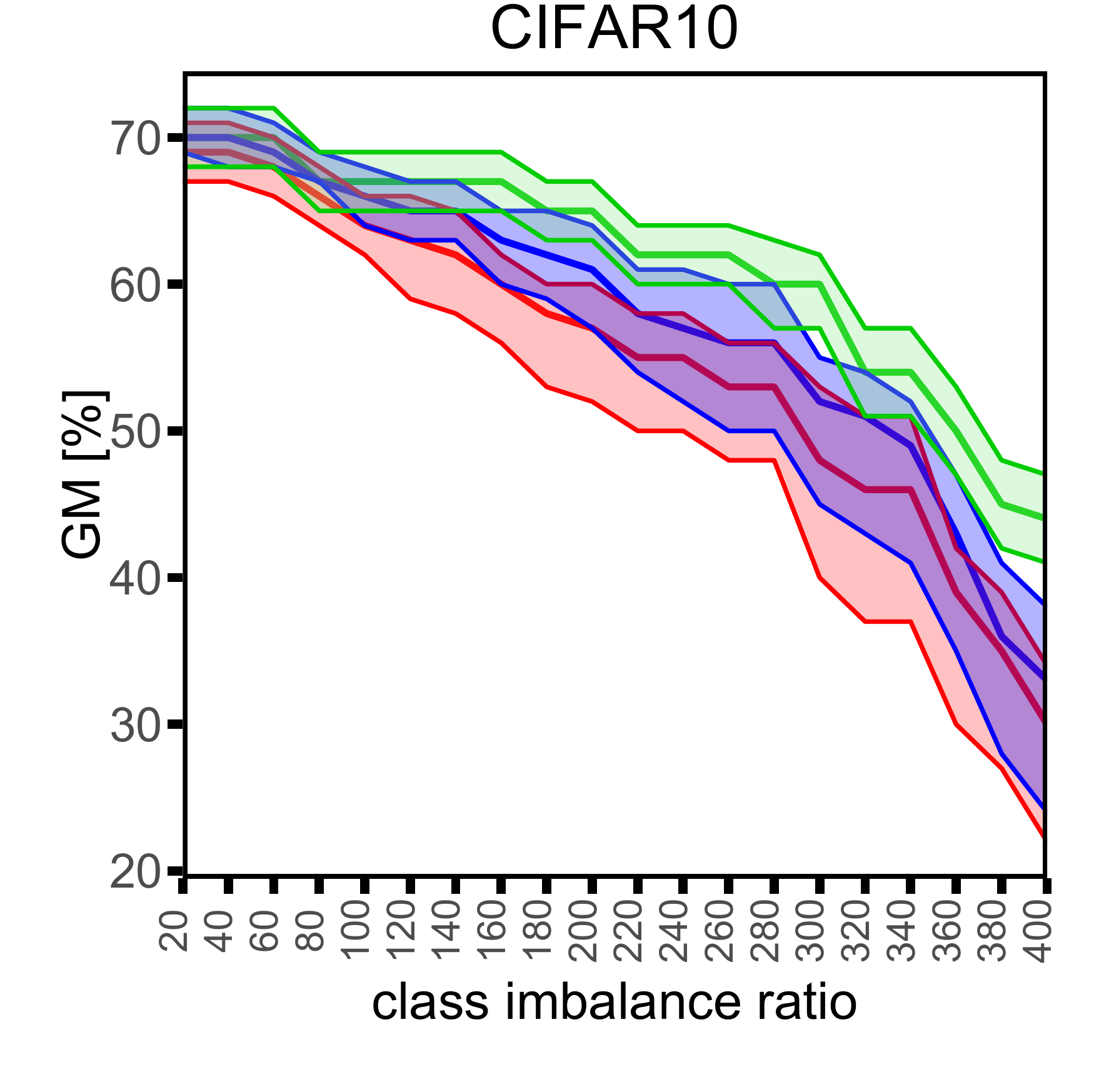}
  \includegraphics[width=0.19\textwidth]{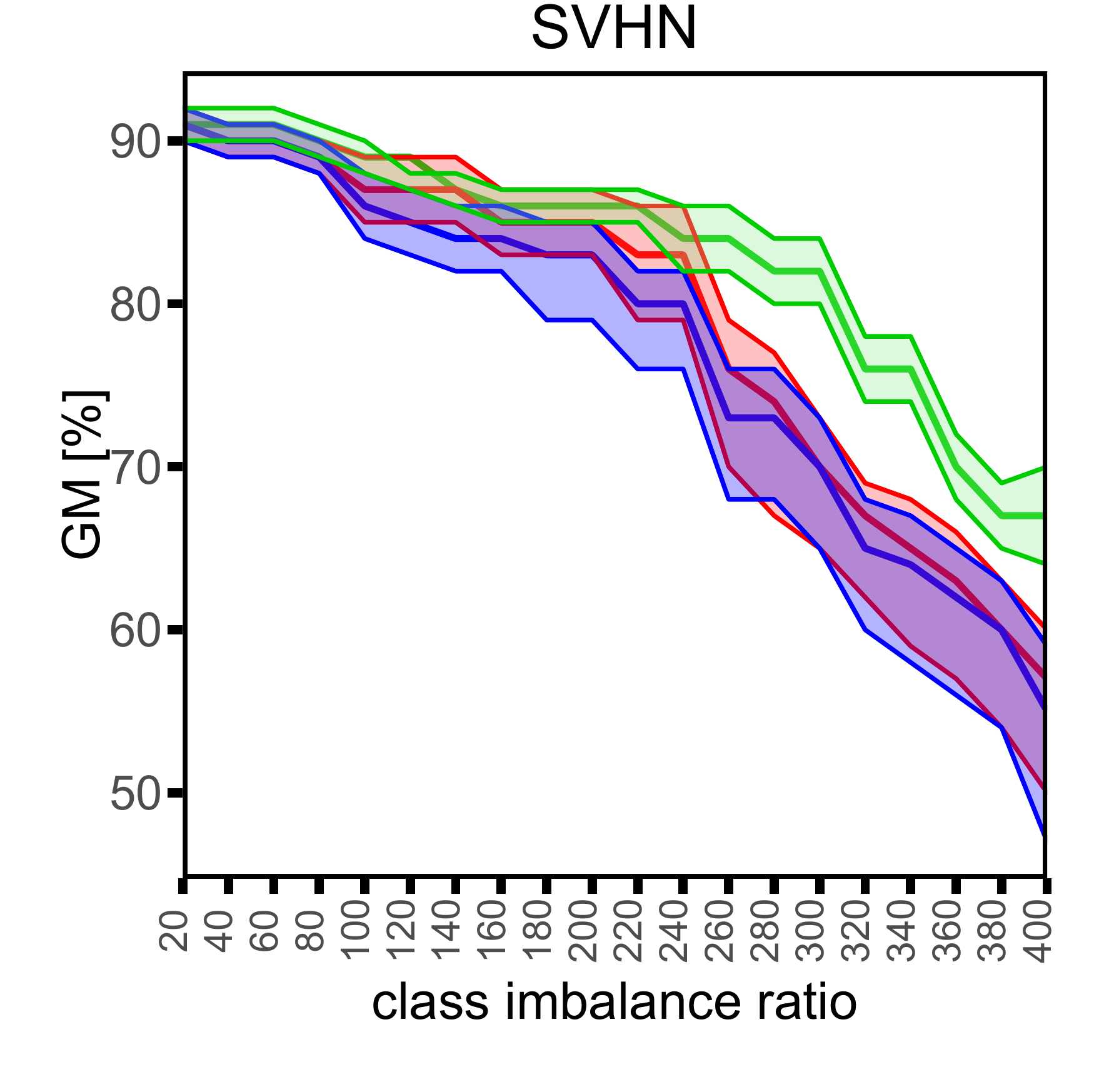}
  \includegraphics[width=0.19\textwidth]{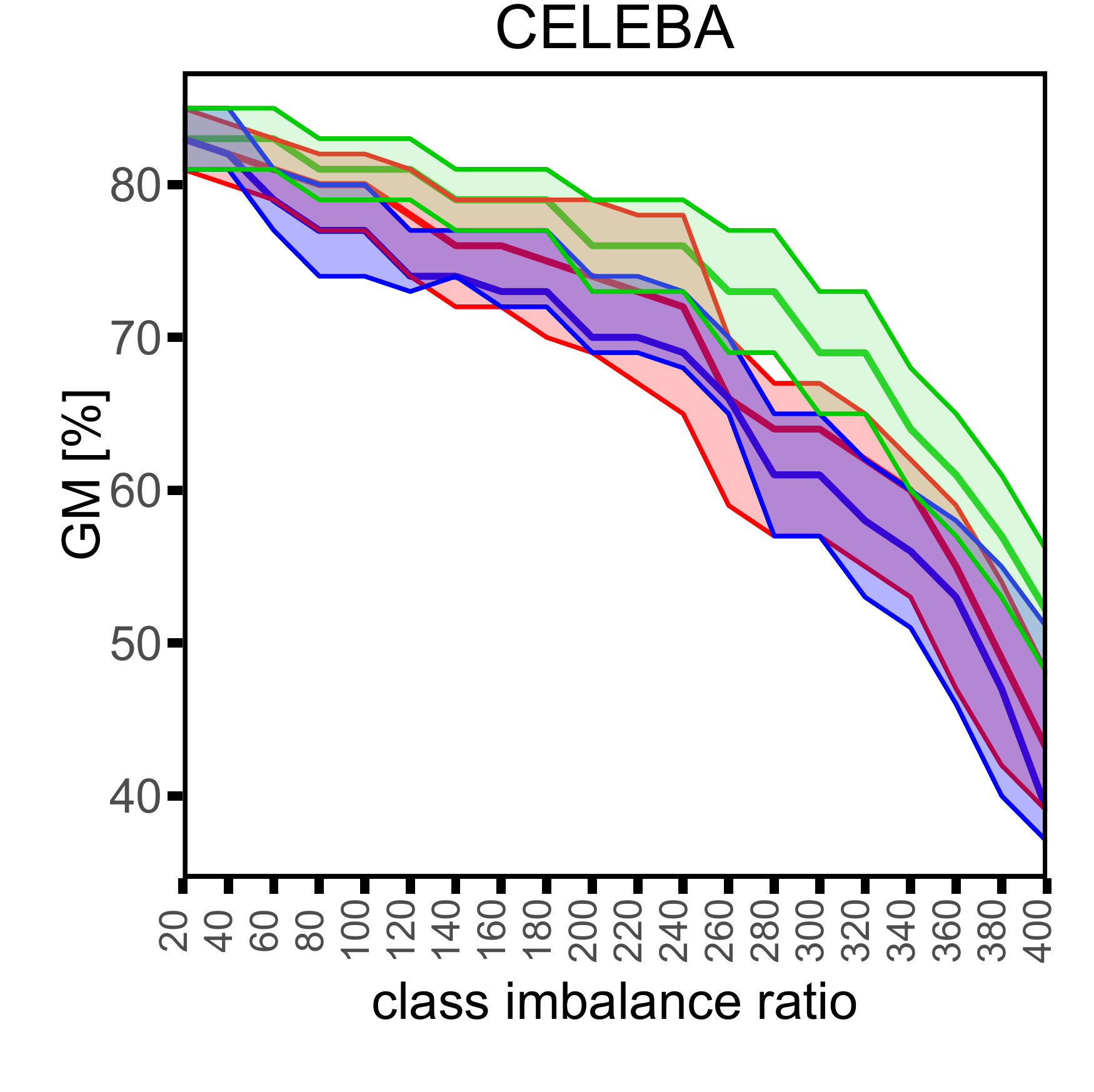}
 
  \includegraphics[width=0.19\textwidth]{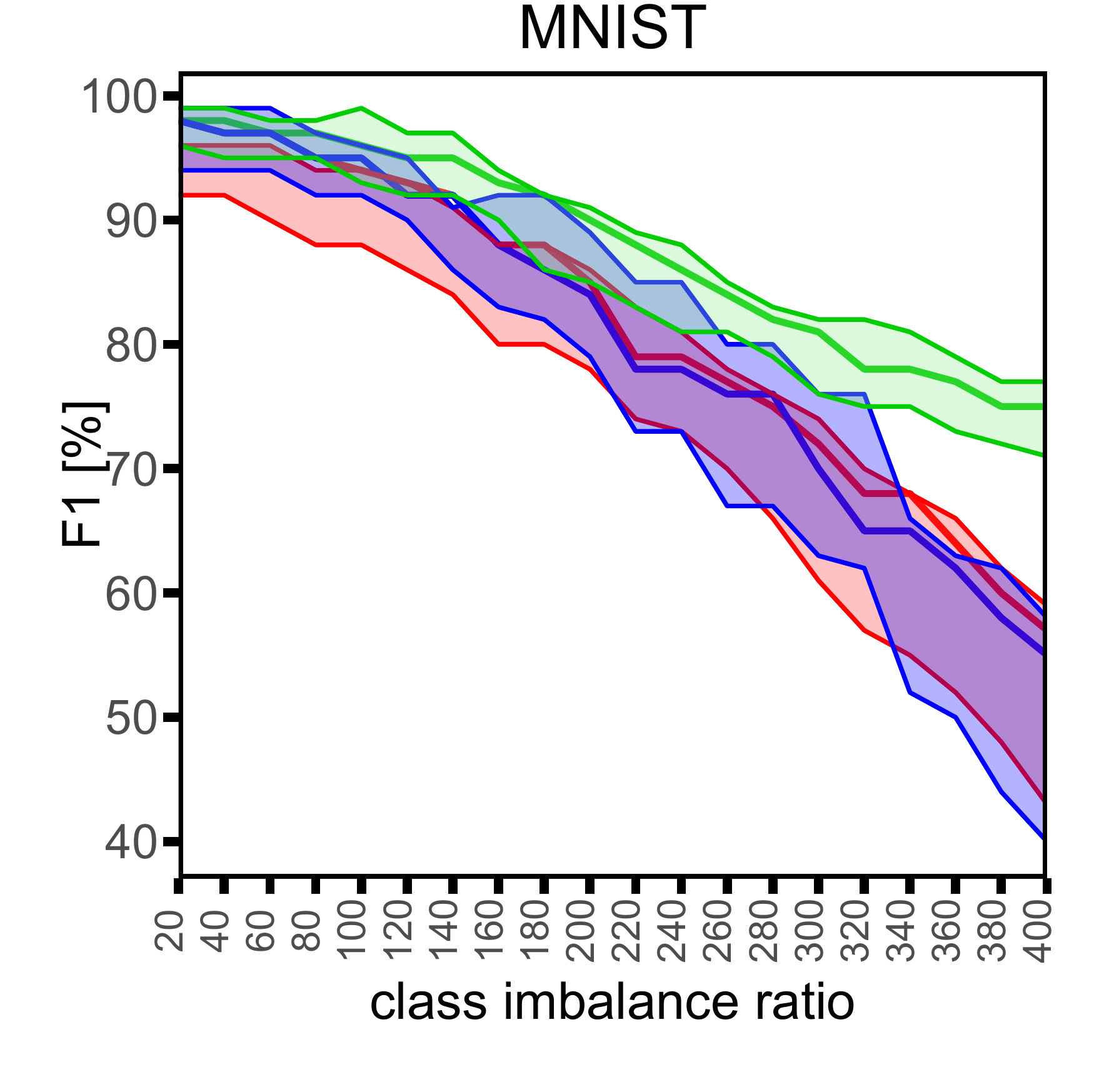}
  \includegraphics[width=0.19\textwidth]{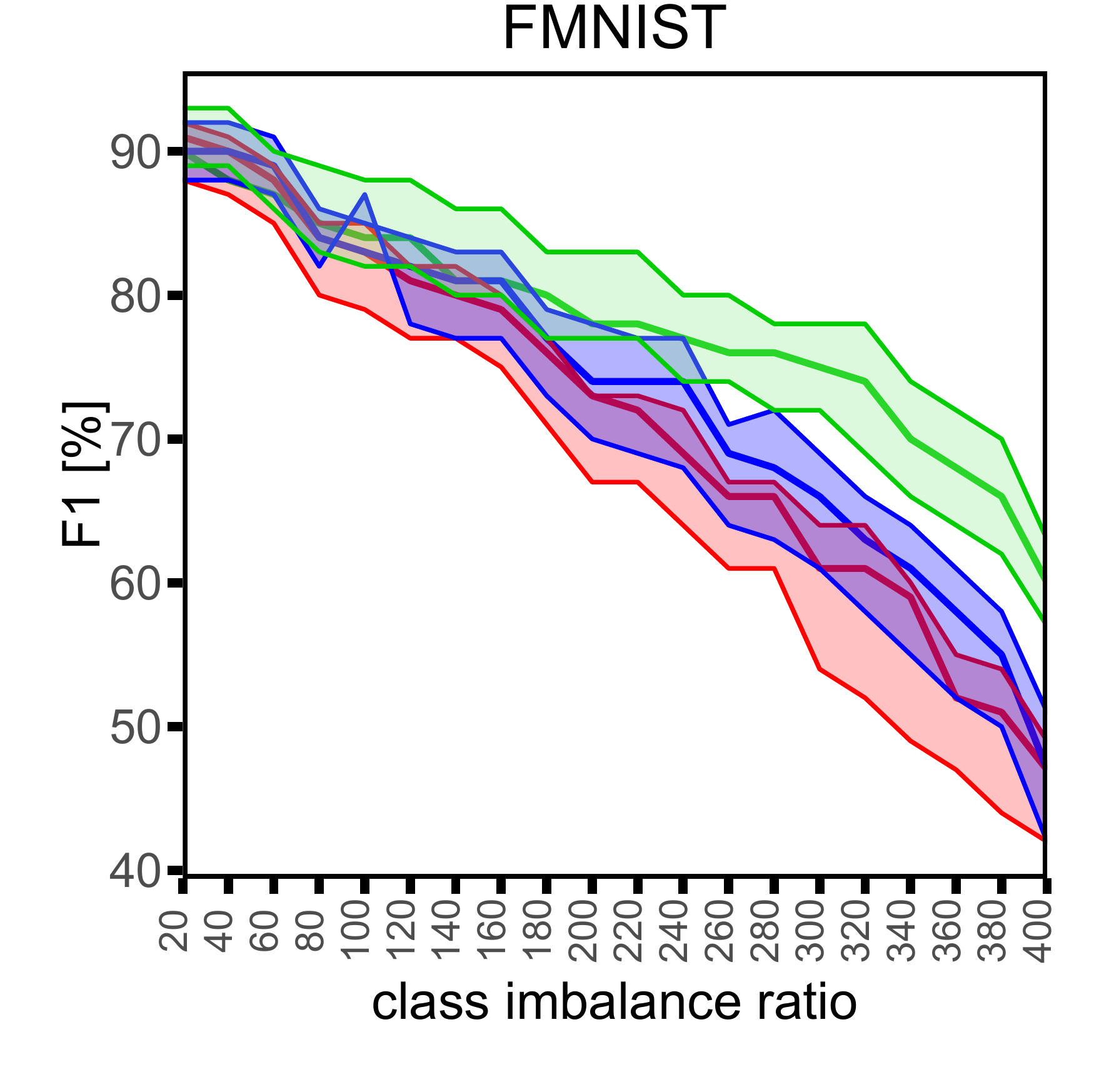}
  \includegraphics[width=0.19\textwidth]{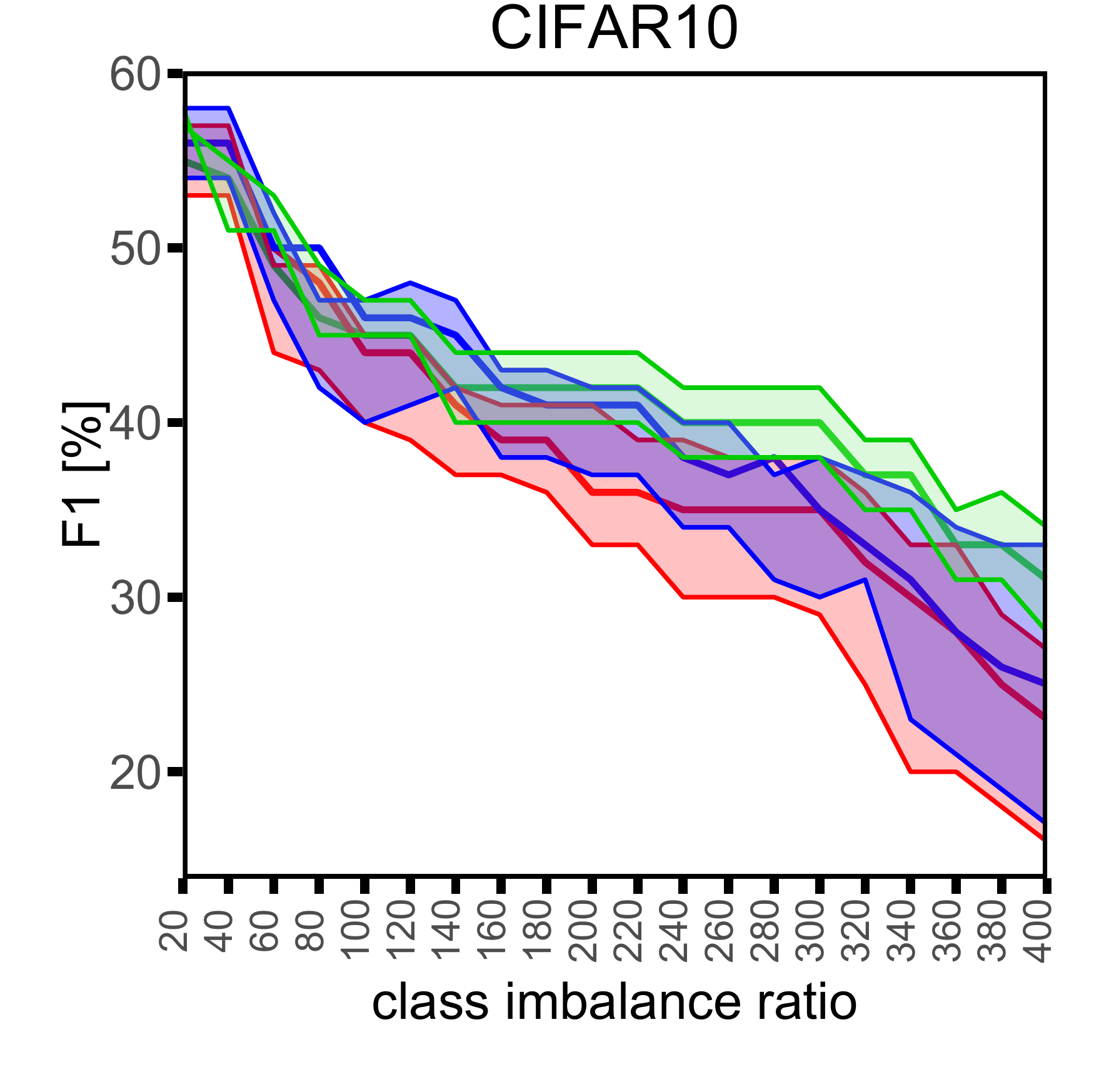}
  \includegraphics[width=0.19\textwidth]{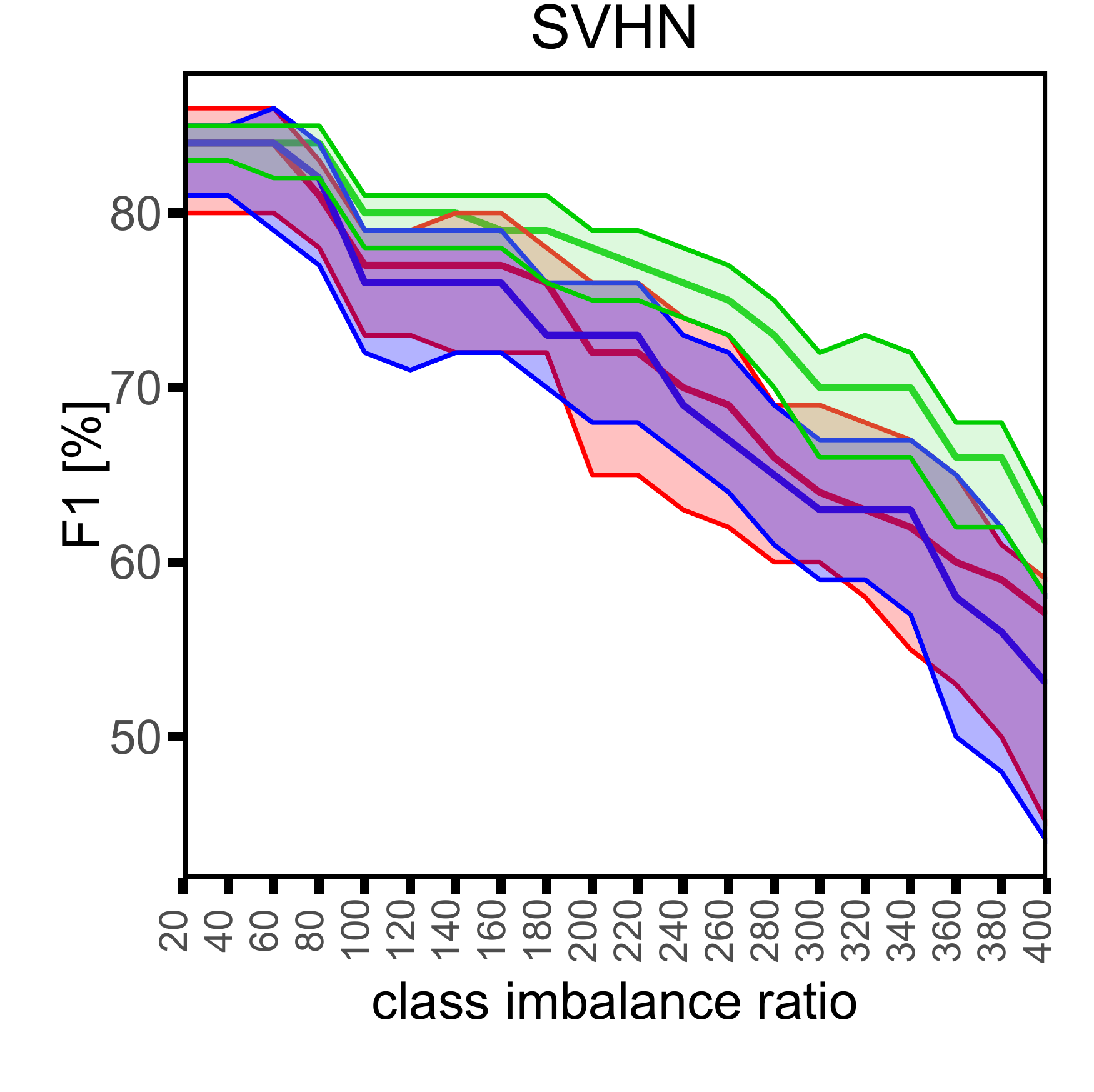}
  \includegraphics[width=0.19\textwidth]{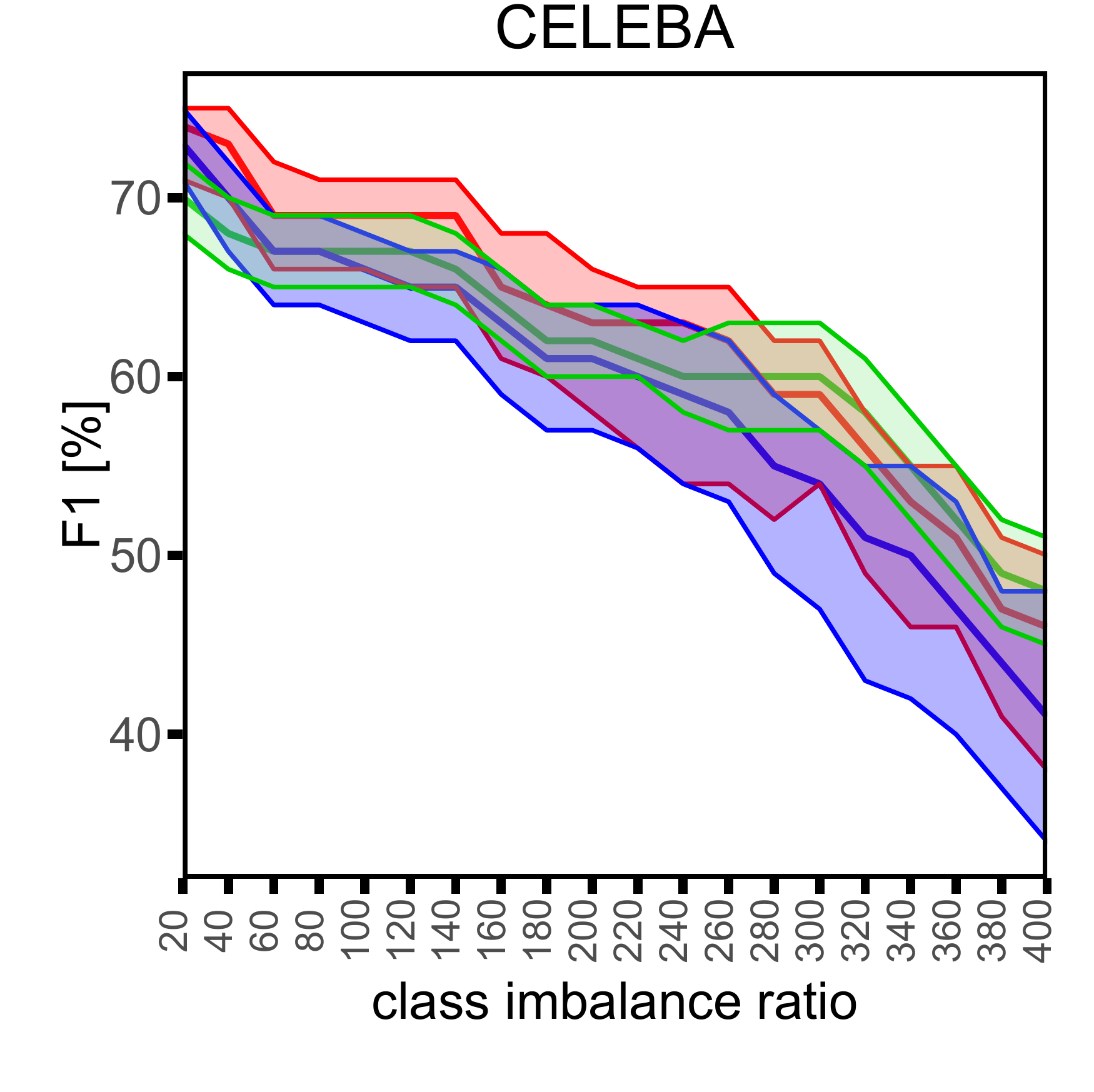}
  \caption{Relationship between imbalance ratio and model stability (expressed as metric spread) for DeepSMOTE and GAN-based models obtained from 20 repetitions of 5-fold CV}
  \label{fig:sta}
\end{figure*}

\noindent \textbf{Robustness to varying imbalance ratios.}  One of the most challenging aspects of learning from imbalanced data lies in creating robust algorithms that can manage various data-level difficulties. Many existing resampling methods return very good results only under specific conditions or under a narrow range of imbalance ratios. Therefore, in order to obtain a complete picture of the performance of DeepSMOTE, we analyze its robustness to varying imbalance ratios in the range of [20,400]. Figure~\ref{fig:rob} depicts the relationship between the three performance metrics and increasing imbalance ratio on five used benchmarks. This experiment allows us not only to evaluate DeepSMOTE and the reference methods under various skewed scenarios, but also offers a bird-eye view on the characteristics of the performance curves displayed by each examined resampling method. An ideal resampling algorithm should be characterized by a high robustness to increasing imbalance ratios, display stable, or small, performance degradation with increased class disproportions. Sharp and significant performance declines indicate breaking points for resampling methods and show when a given algorithm stops being capable of generating useful instances and countering class imbalance. 

\noindent Analyzing Figure~\ref{fig:rob} allows us to draw several interesting conclusions. First, Experiment 1 shows that pixel-based solutions are inferior to their GAN-based counterparts. However, we can see that this observation does not hold for extreme values of imbalance ratios. When the disproportion among classes increases, pixels-based methods (especially MC-CCR and MC-RBO) start displaying increased robustness. On the contrary, the two GAN-based methods are more sensitive to an increased imbalance ratio and we can observe a more rapid decline in their predictive power. This can be explained by two factors: the method by which resampling approaches use the original instances and the issue of small sample size. The former factor shows the limitations of GAN-based methods. While they focus on instance generation and creating high-quality images, they do not possess more sophisticated mechanisms on where to precisely inject new artificial instances. With higher imbalance ratios, this placement starts playing a crucial role, as the classifier needs to handle more and more difficult bias. Current GAN-based models use relatively simplistic mechanisms for this issue. On the contrary, pixel-based methods rely on more sophisticated mechanisms, (e.g., MC-CCR uses an energy-based function, while MC-RBO uses local optimization for positioning their artificial instances). With increasing imbalance ratios, such mechanisms start to dominate simpler GAN-based solutions, making pixel-based approaches more robust to extreme imbalance ratios. The latter factor of small sample size also strongly affects GAN-based algorithms. With extreme imbalance, we have less and less minority instances at our disposal, making it more difficult to train effective GANs. 

\noindent Compared to both pixel-based and GAN-based approaches, DeepSMOTE displays an excellent robustness even to the highest imbalance ratios. We can see that DeepSMOTE is able to effectively handle such a challenging scenario, displaying the lowest decline of performance on all evaluated metrics. This can be attributed to the fact that SMOTE generates artificial instances following class geometry, while using only nearest neighbors for instance generation. Hence, DeepSMOTE is not affected as strongly as GAN-based approaches by a small sample size andthe  need for smart placement of artificial instances, leading to excellent robustness (\textbf{RQ5 answered}).

\smallskip
\noindent \textbf{Model stability under varying imbalance ratios.} Another important aspect of evaluating modern resampling algorithms is their stability. We need to evaluate how a given model reacts to small perturbations in data, as we want to evaluate its generalization capabilities. Models that display high variance under such small changes cannot be treated as stable and thus should not be preferred. It is especially crucial in the learning from imbalanced data area, as we want to select a resampling algorithm that will generate information-rich artificial instances under any data permutations. 

\noindent In order to evaluate this, we have measured the spread of performance metrics for DeepSMOTE and GAN-based algorithms under 20 repetitions of 5-fold cross validation. During each CV repetition, minority classes were created randomly from the original balanced benchmarks. This ensured that we not only measure the stability to training data permutation within a single dataset instance, but we also measure the possibility of creating minority classes with instances of varying difficulties. Figure~\ref{fig:sta} shows the plots of three resampling methods with shaded regions denoting the variance of results. GAN-based approaches display increasing variance under higher imbalance ratios, showing that those approaches cannot be considered as stable models for challenging imbalanced data problems. DeepSMOTE returned the lowest variance within those metrics, showcasing the high stability of our resampling algorithm. This information enriches our previous observation regarding the robustness of DeepSMOTE. Joint analysis of Figures~\ref{fig:rob} and~\ref{fig:sta} allows us to conclude that DeepSMOTE can handle extreme imbalance among classes, while generating stable models under challenging conditions (\textbf{RQ6 answered}).

 \section{Discussion}
 \label{sec:les}

\begin{itemize}
\item \textbf{Simple design is effective.} DeepSMOTE is an effective approach for countering class imbalance and training skew-insensitive deep learning classifiers. It outperforms state-of-the-art solutions, and is able to work on raw image representations. DeepSMOTE is composed of three components: an encoder/decoder is combined with a dedicated loss function and SMOTE-based resampling. This simplicity makes it an easy to understand, transparent, yet very powerful method for handling class imbalance in deep learning. 
\item \textbf{Dedicated data encoding for artificial instance generation.} DeepSMOTE uses a two-phase approach that first trains a dedicated encoder/decoder architecture and then uses it to obtain a high quality embedding for the oversampling procedure. This allows us to find the best possible data representations for oversampling, allowing SMOTE-based generation to enrich the training set of minority classes. 
\item \textbf{Effective placement of artificial instances.} DeepSMOTE follows the geometric properties of minority classes, creating artificial instances on borders among classes. This leads to improved training of discriminative models on datasets balanced with DeepSMOTE. 
\item \textbf{Superiority over pixel-based and GAN-based algorithms.} DeepSMOTE outperforms state-of-the-art resampling approaches. By being able to work on raw images and extracting features from them, DeepSMOTE can generate more meaningful artificial instances than pixel-based approaches, even while using relatively simpler rules for instance generation. By using efficient and dedicated data embeddings, DeepSMOTE can better enrich minority classes under varying imbalance ratios than GAN-based solutions.
\item \textbf{Easy to use.} One of the reasons behind the tremendous success of the original SMOTE algorithm was its easy and intuitive usage. DeepSMOTE follows these steps, as it is not only accurate, but also an attractive off-the-shelf solution.  Our method is easy to tune and use on any data, both as a black-box solution and as a steppingstone for developing novel and robust deep learning architectures. As deep learning is being used by a wider and wider interdisciplinary audience, such a characteristic is highly sought after.
\item \textbf{High quality of generated images.} DeepSMOTE can return high quality artificial images that under visual inspection do not differ from real ones. This makes DeepSMOTE an all-around approach, since the generated images are both sharp and information rich. 
\item \textbf{Excellent robustness and stability.} DeepSMOTE can handle extreme imbalance ratios, while being robust to small sample size and within-data variance. DeepSMOTE is less prone to variations in training data than any of the reference methods. It is a stable oversampling approach that is suitable for enhancing deep learning models deployed in real-world applications. 
\end{itemize}

\section{Conclusion}
\label{sec:con}

\noindent \textbf{Summary.} We proposed DeepSMOTE, a novel and transformative model that fuses the highly popular SMOTE algorithm with deep learning methods. This allows us to create an efficient oversampling solution for training deep architectures on imbalanced data distributions. DeepSMOTE can be seen as a data-level solution to class imbalance, as it creates artificial instances that balance the training set, which can then be used to train any deep classifier without suffering from bias. DeepSMOTE uniquely fulfilled three crucial characteristics of a successful resampling algorithm in this domain: the ability to operate on raw images, creation of efficient low-dimensional embeddings, and the  generation of high-quality artificial images. This was made possible by our novel architecture that combined an encoder/decoder framework with SMOTE-based oversampling and an  enhanced loss function. Extensive experimental studies show that DeepSMOTE not only outperforms state-of-the-art pixel-based and GAN-based oversampling algorithms, but also offers unparalleled robustness to varying imbalance ratios with high model stability, while generating artificial images of excellent quality. 

\smallskip
\noindent \textbf{Future work.} Our next efforts will focus on enhancing DeepSMOTE with information regarding class-level and instance-level difficulties, which will allow it to better tackle challenging regions of the feature space. We plan to enhance our dedicated loss function with instance-level penalties for focusing the encoder/decoder training on instances that display borderline / overlapping characteristics, while discarding outliers and noisy instances. Such a compound skew-insensitive loss function will bridge the worlds between data-level and algorithm-level approaches to learning from imbalanced data. Furthermore, we want to make DeepSMOTE suitable for continual and lifelong learning scenarios, where there is a need for handling dynamic class ratios and generating new artificial instances. We envision that DeepSMOTE may not only help to counter online class imbalance, but also help increase the robustness of lifelong learning models to catastrophic forgetting. Finally, we plan to extend DeepSMOTE to incorporate other data modalities, such as graphs. 




\bibliographystyle{IEEEtran}
\bibliography{dsmrefs}

\end{document}